\documentclass[default,iicol]{sn-jnl}

\usepackage[english]{babel}
\usepackage{amsmath}
\usepackage{graphicx}
\usepackage{multirow}
\usepackage[normalem]{ulem}
\usepackage[export]{adjustbox}
\usepackage{subfig}
\jyear{2023}%

\begin{document}

\title[Article Title]{U-DIADS-Bib: a full and few-shot pixel-precise dataset for document layout analysis of ancient manuscripts}


\author*[1]{\sur{Silvia Zottin}}
\email{zottin.silvia@spes.uniud.it}

\author[1,2]{\sur{Axel {De Nardin}}}
\email{denardin.axel@spes.uniud.it}

\author[3]{\sur{Emanuela Colombi}}
\email{emanuela.colombi@uniud.it}

\author[1]{\sur{Claudio Piciarelli}}
\email{claudio.piciarelli@uniud.it}

\author[3]{\sur{Filippo Pavan}}
\email{pavan.filippo@spes.uniud.it}

\author[1]{\sur{Gian Luca Foresti}}
\email{gianluca.foresti@uniud.it}

\affil[1]{\orgdiv{Department of Mathematics, Computer Science and Physics}, \orgname{University of Udine}, \orgaddress{Udine}, \country{Italy}}
\affil[2]{\orgdiv{Department of Engineering and Architecture}, \orgname{University of Trieste}, \orgaddress{Trieste}, \country{Italy}}
\affil[3]{\orgdiv{Department of Humanities and Cultural Heritage}, \orgname{University of Udine}, \orgaddress{Udine}, \country{Italy}}


\abstract{
Document Layout Analysis, which is the task of identifying different semantic regions inside of a document page, is a subject of great interest for both computer scientists and humanities scholars as it represents a fundamental step towards further analysis tasks for the former and a powerful tool to improve and facilitate the study of the documents for the latter. However, many of the works currently present in the literature, especially when it comes to the available datasets, fail to meet the needs of both worlds and, in particular, tend to lean towards the needs and common practices of the computer science side, leading to resources that are not representative of the humanities real needs. 
For this reason, the present paper introduces U-DIADS-Bib, a novel, pixel-precise, non-overlapping and noiseless document layout analysis dataset developed in close collaboration between specialists in the fields of computer vision and humanities. 
Furthermore, we propose a novel, computer-aided, segmentation pipeline in order to alleviate the burden represented by the time-consuming process of manual annotation, necessary for the generation of the ground truth segmentation maps.
Finally, we present a standardized few-shot version of the dataset (U-DIADS-BibFS), with the aim of encouraging the development of models and solutions able to address this task with as few samples as possible, which would allow for more effective use in a real-world scenario, where collecting a large number of segmentations is not always feasible.
}

\keywords{Document Segmentation, Document Layout Analysis, Document Image Analysis, Document Image Dataset, Pixel-level Annotation, Few-Shot Learning}



\maketitle

\section{Introduction}\label{sec1}
In recent years, interest in cultural heritage is also spreading within the IT community. The cooperation between computer scientists and humanists is becoming more and more prominent, steadily leading to a better understanding of this type of data for the former and a simplification in the work of the latter.
One of the activities that humanists are particularly interested in, and usually carry out as part of the process to study an ancient manuscript, is to identify the logical and physical structure of the documents and their organization, highlighting and extracting specific sets of elements in order to achieve a higher level of understanding  regarding their contents and the correlations with other manuscripts of the same kind.
In the IT field, this activity is known as page segmentation which, together with text-line segmentation and baseline detection, constitutes the macro area referred to as document layout analysis. 

However, to be able to address these kinds of problems, especially when working with Machine Learning models, we need to access a substantial amount of data regarding the manuscripts we want to analyze and, in particular, we need the Ground Truth (GT) that represent the target of the segmentation we want to obtain. GT is essential not only to train the models but also to evaluate their performance in a quantitative fashion which, in turn, provides a reliable way of comparing different approaches in a standardized what and assessing if the developed models are accurate enough to be employed in real contexts to help humanists in their work.

For GT maps to be suitable for the task at hand, they must be accurate and consequently, the segmentation annotation must be as precise as possible down to the pixel-level. This operation is very difficult mainly for two reasons. The first is that creating GTs takes a long time and requires domain-specific knowledge, which only an expert humanist possesses. The second is that when analyzing images of ancient manuscripts we have to take into account several factors that make the appearance of the different pages non-homogeneous. These include heterogeneous degrees of degradation of the pages, a non-uniform layout, ink stains, scratches, show-through, and out-of-calibration scanning. Therefore, any automation of segmentation activity introduces a lot of noise into the GTs, invalidating the results and the outputs.

In addition, in the current literature, there are many examples of datasets that address the page segmentation task, however in all of them the segmentation classes are either not detailed enough, distinguishing only between foreground and background~\cite{Chinese}, or including only a small number of layout classes~\cite{diva} or they provide relatively low quality and noise segmentation maps~\cite{HBA}. 
Finally, the datasets currently available in the literature that address the page segmentation task are mono-alphabetic, presenting only one type of grapheme such as only Latin alphabet~\cite{diva}, only Arabic alphabet~\cite{RASM2018} or only Chinese alphabet~\cite{Chinese}.

For these reasons, in this work, we present the Uniud - Document Image Analysis DataSet - Bible version (U-DIADS-Bib), a dataset for document layout segmentation developed in close collaboration between specialists in the fields of computer vision and humanities studies. The latter analyzed the selected manuscripts and agreed on which aspects were relevant for studying them, providing valuable insight regarding the choice of the segmentation classes to be selected for the dataset. The computer vision specialists, on the other hand, ensured that the produced GTs were consistent across the instances of the same dataset and that they also presented the characteristics which define a high quality modern semantic segmentation-oriented dataset such as the absence of overlap between the classes, precision to the pixel level and absence of noise. 
All these characteristics represent necessary conditions for a meaningful analysis of the available data via Machine and Deep Learning approaches, however, none of the datasets currently available in the literature satisfies all of them at the same time.
The result of this process led to a selection of six semantic classes which are namely: background, main text, paratext, decoration, title and chapter headings.
Furthermore, the manuscripts that have been selected present the Latin alphabet and the Abjad consonantal alphabet, in particular the Syriac one, which therefore have very different graphenes, making the page segmentation task more challenging.


A significant challenge emerged at the beginning of this collaboration has been represented by the significant amount of time needed to perform the manual segmentation process necessary to obtain the GT segmentation maps for all the images characterizing the dataset.
For this reason, in addition to the dataset presentation, we also propose a segmentation pipeline that reduces the workload for the humanities experts by alternating their manual segmentation process with the one produced by a Machine Learning model trained on just a few samples. The segmentations thus obtained are then used as a coarse baseline on which the humanities experts perform a refinement process to obtain the final ground truth segmentation maps. This process has proven to be considerably less time-consuming then performing the segmentation from scratch.

Furthermore, in relation to the aforementioned problem, we believe that developing Deep Learning models which are able to learn an effective representation of the data by starting from just a few samples is a key step in the context of document layout analysis as it will allow to reduce the need of segmented instances for real-world applications. For this reason, we also present a standardized few-shot version of the dataset which we believe will help stimulate more efforts toward this goal.
Finally, a benchmark obtained by applying a set of popular semantic segmentation models to the two versions of our dataset is provided as a baseline for future studies.

The remainder of the paper is organized as follows. Section~\ref{sec2} reviews historical handwritten datasets available in the current literature. A detailed description of the U-DIADS-Bib dataset and of the process of making ground truth is given in  Section~\ref{sec3}. Section~\ref{sec4} defines the experimental results of the whole dataset and the few-shot dataset. Finally, Section~\ref{sec5} highlights conclusions and future works.

\begin{figure*}[t]%
    \centering
    \subfloat[\centering Latin 2, page 232]{{\includegraphics[width=.205\linewidth]{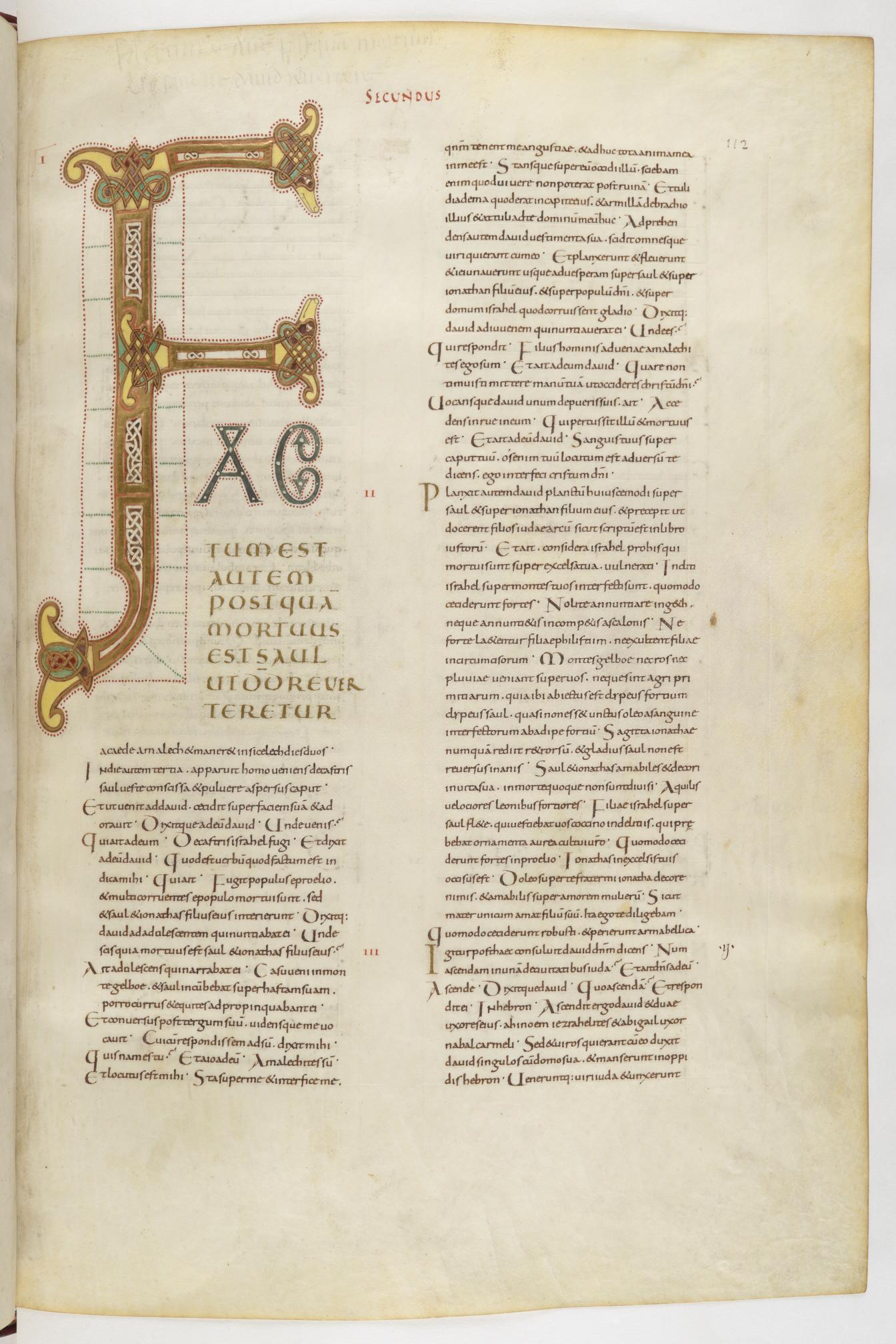} \label{fig:2page} }}%
    \qquad
    \subfloat[\centering Latin 14396, page 251]{{\includegraphics[width=.205\linewidth]{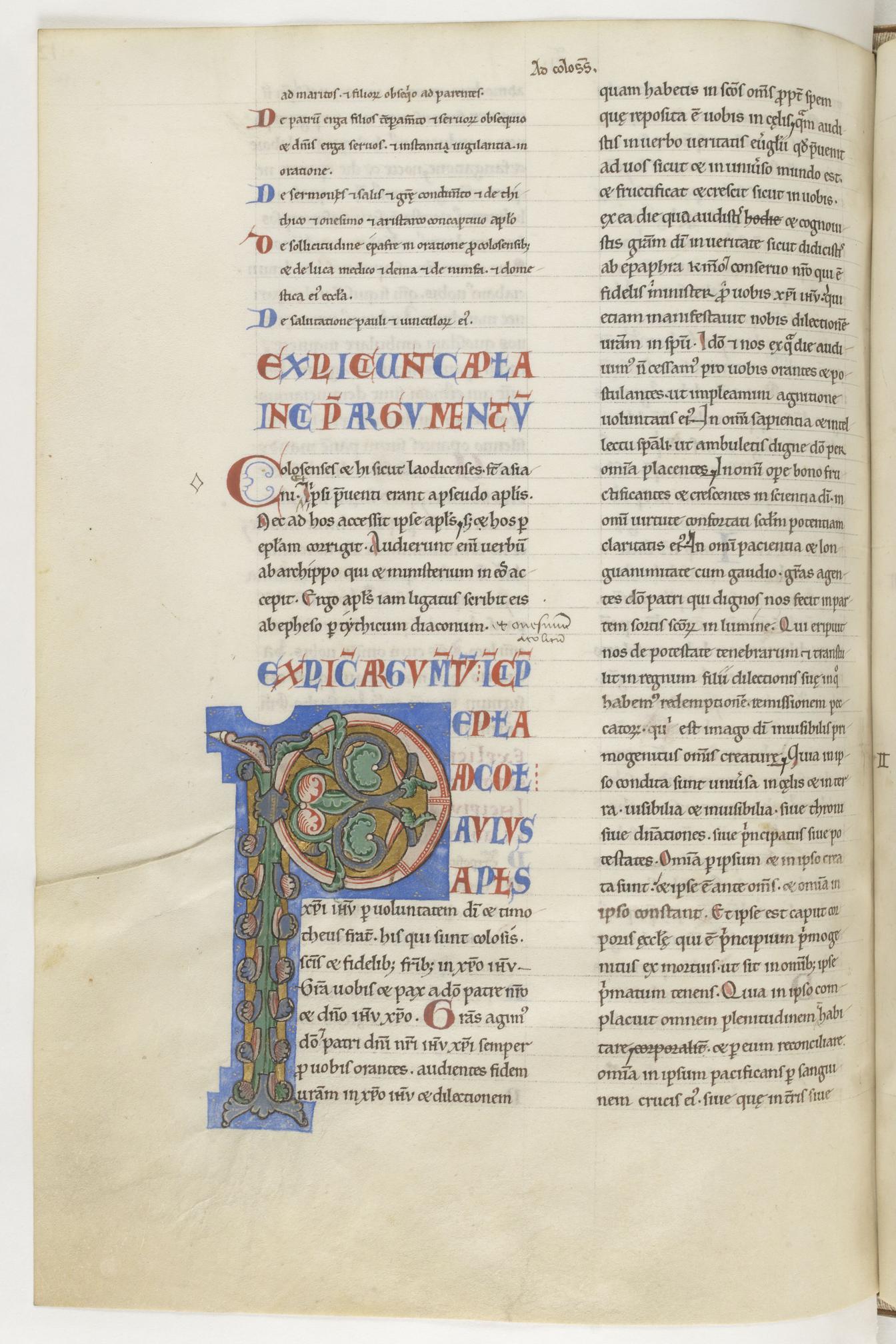} \label{fig:14396page}}}
    \qquad
    \subfloat[\centering Latin 16746, page 255]{{\includegraphics[width=.205\linewidth]{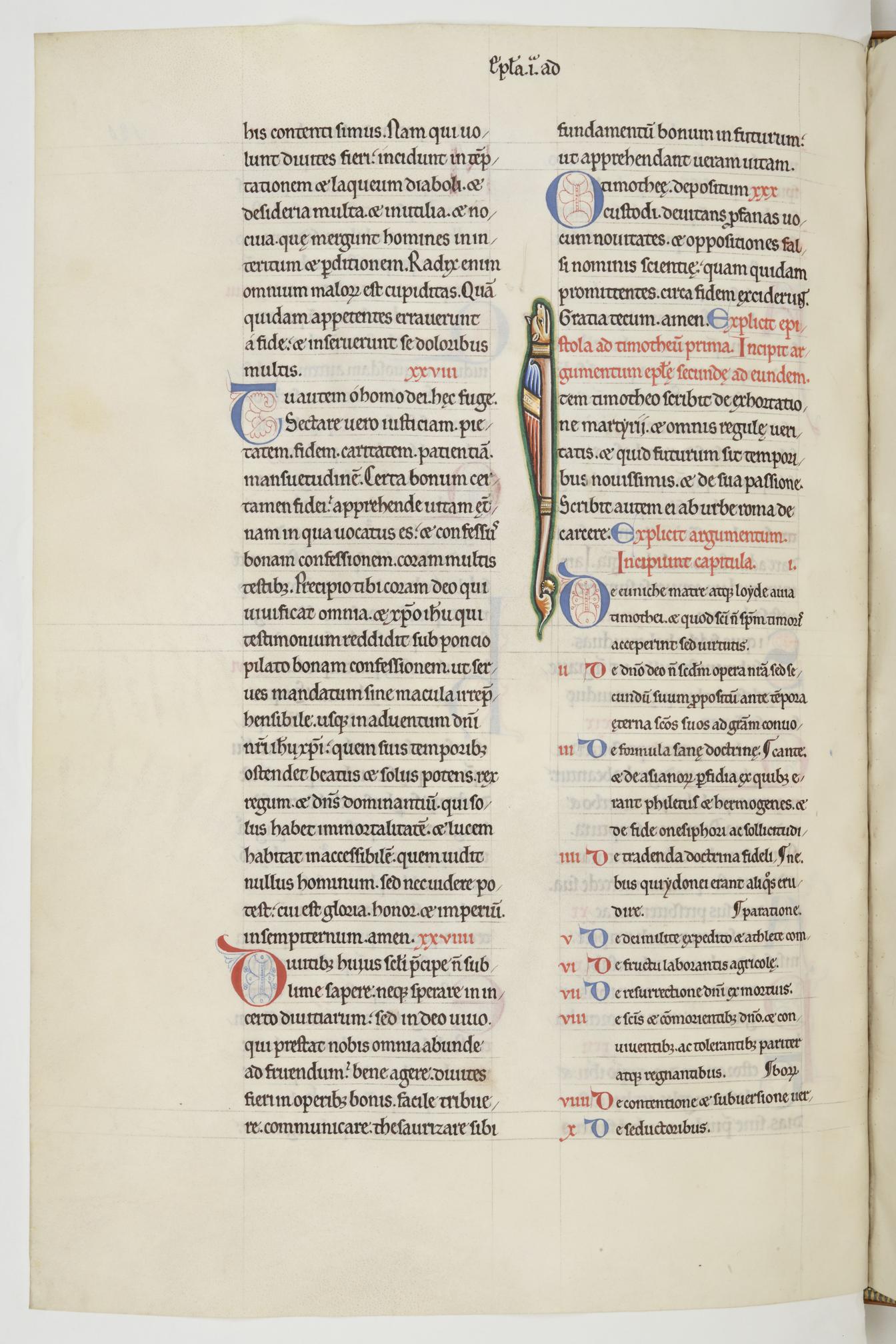}  \label{fig:16746page}}}%
    \qquad
    \subfloat[\centering Syriaque 341, page 53]{{\includegraphics[width=.205\linewidth]{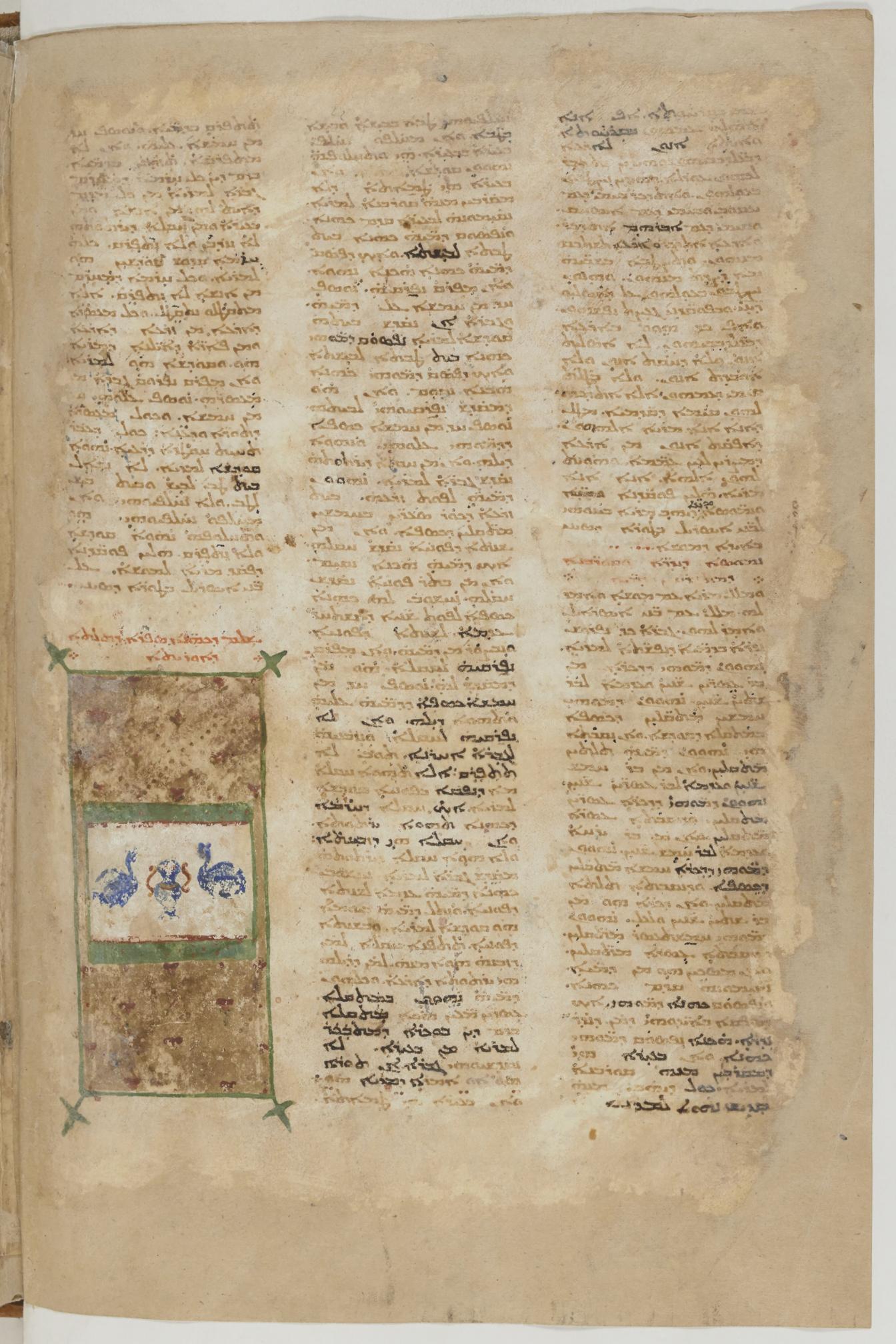}  \label{fig:341page}}}%
    \qquad
    \subfloat[\centering Latin 2 detail]{{\includegraphics[width=.205\linewidth]{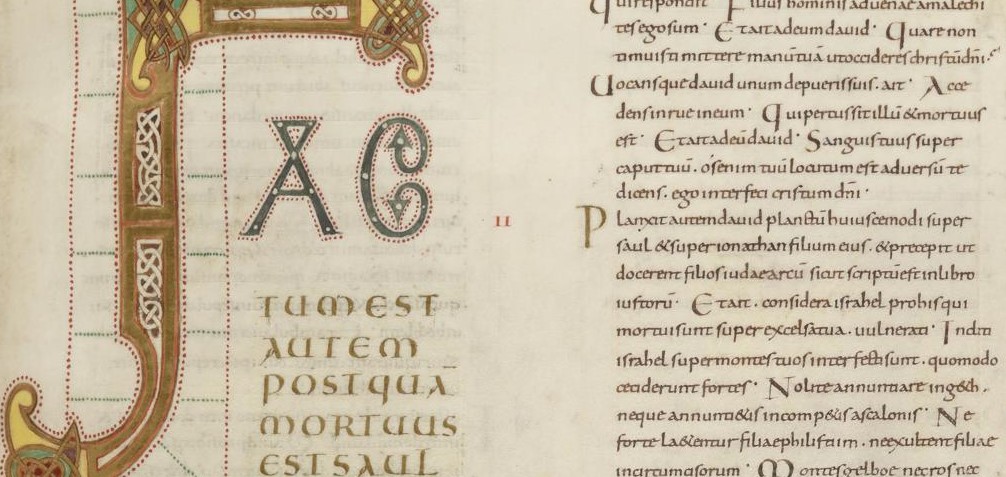} \label{fig:2det}}}%
    \qquad
    \subfloat[\centering Latin 14396 detail]{{\includegraphics[width=.205\linewidth]{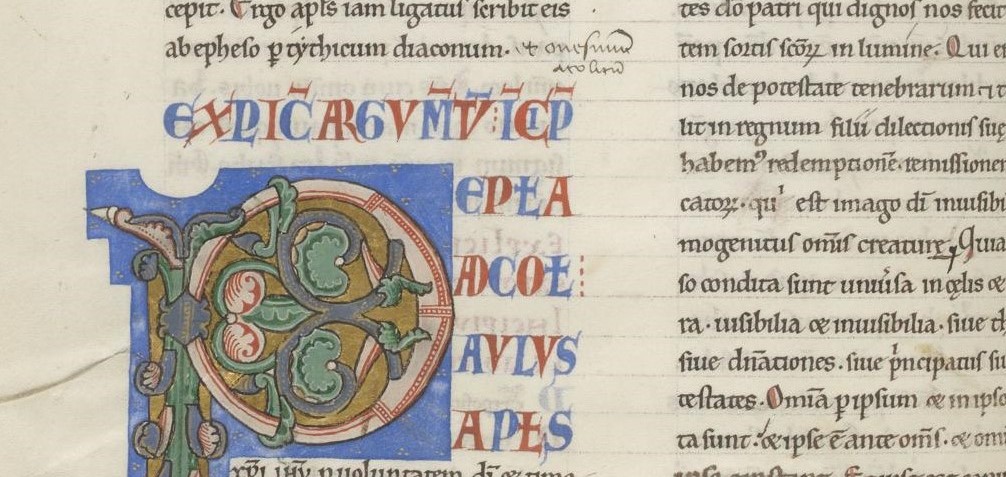} \label{fig:14396det}}}
    \qquad
    \subfloat[\centering Latin 16746 detail]{{\includegraphics[width=.205\linewidth]{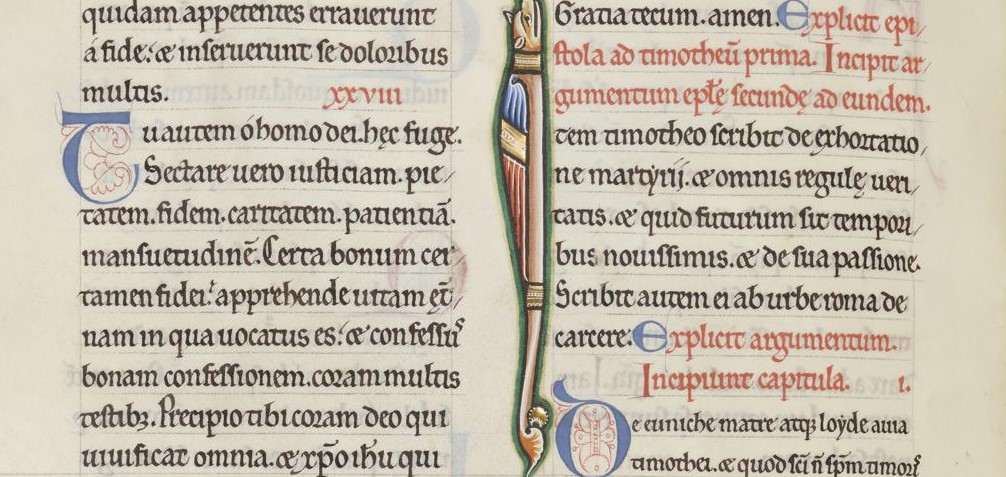} \label{fig:16746det}}}
    \qquad
    \subfloat[\centering Syriaque 341 detail]{{\includegraphics[width=.205\linewidth]{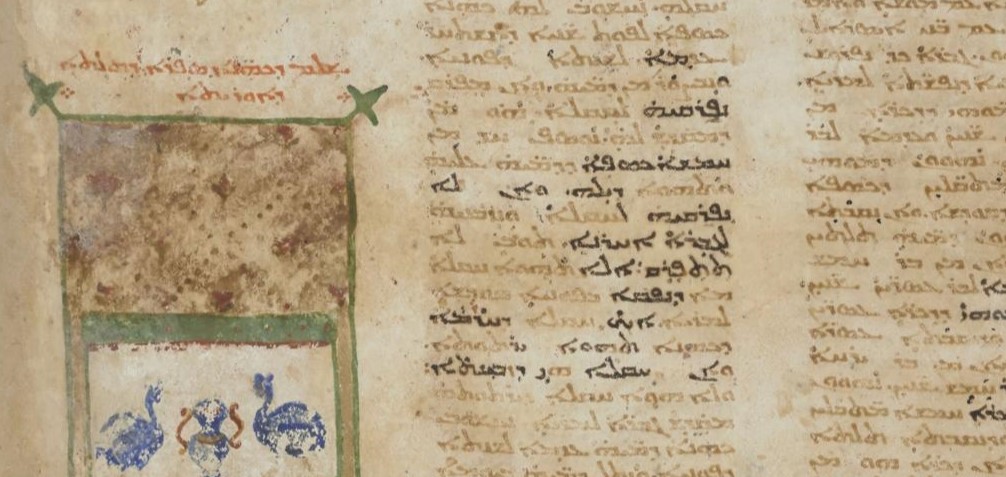} \label{fig:341det}}}%
    \caption{Samples from the 4 manuscripts (Latin 2, Latin 14396, Latin 16746 and Syriaque 341) presents in U-DIADS-Bib. Fig.~\ref{fig:2page}\textendash~\ref{fig:341page} show a full page for each manuscripts, while Fig.~\ref{fig:2det}\textendash~\ref{fig:341det} show a detail extracted from each of them.}%
    \label{fig:examplepage}%
\end{figure*}

\section{Related works}\label{sec2}

A systematic literature review of existing  datasets for historical document image analysis is presented in~\cite{Nikolaidou2022}.
In general, there are various datasets for historical manuscript image analysis for different tasks (e.g. baseline detection~\cite{Gruning, Kiessling}, text-line segmentation~\cite{Potanin, Gatos}, handwriting recognition~\cite{Fischer, Wüthrich, Dolfing, Kassis, Fischer1}, writer identification~\cite{Adam2018, Cilia, Fiel}), but datasets for layout analysis with pixel-precision segmentation are very limited.

In the literature for layout analysis, there are two types of datasets that differ in the type of ground truth they make available. 
The first type provides the ground truth for the different segmentation classes in the form of bounding boxes. These can be rectangles or polygons that provide a coarse representation of the structure of a document page and are often stored in XML format.
RASM2018~\cite{RASM2018} is one of the datasets for layout analysis of this type. This dataset presents a collection of 100 images of historical scientific manuscripts with text in the Arabic language. 
Another is the Pinkas~\cite{Pinkas} dataset, which consists of 30 images from medieval Hebrew manuscripts. Page segmentation of this dataset determines the main text, side text, signature marks and dates.
Finnish Court Records-sub500~\cite{quiros} is a 500-page of Finnish notarial records. The dataset is annotated at the image level using six different region types such as page number, paragraph and table.
HisClima~\cite{hisclima} is a database of handwritten weather ship logs and comprises 208 pages with tables and 211 pages with descriptive text. Layout analysis of this dataset is focused on distinguishing between the main text and the table regions.
Horae~\cite{horae} dataset consists of 557 images derived from Books of Hours. This dataset classifies layout areas into a wide set of categories such as text region, miniature, decorated border, illustrated border, decorated initial, music notation and ornamentation.

However, highly edited texts such as ancient manuscripts have complex layouts and the use of polygons to delimit the layout regions does not provide a sufficiently accurate and truthful representation for them. In some cases, there are extensive regions in which multiple layout classes almost overlap with each other making the use of polygons unreliable. The pixel-by-pixel classification solution of the layout is therefore a preferred option in this scenario and represents the second category of datasets used in this context.
PHTD~\cite{phtd} is a 140-page dataset of handwritten documents in the Persian language. This dataset provides only two classes of segmentation: text and background, making it, effectively, a binarization dataset more than a layout segmentation one.
Similar to the previous one, also HDRC-CHINESE~\cite{Chinese} dataset provides the layout analysis with the segmentation of text and non-text (background) classes. This dataset consists of page images derived from 37 different historical Chinese family record books.
Bukhari et al.~\cite{Bukhari} present a dataset consisting of 38 document images of three different Arabic manuscripts. The segmentation occurs into three classes: side-notes text, main-body text and background.
DIVA-HisDB~\cite{diva} consists of three medieval manuscripts with a total of 150 pages annotated at pixel-level for the following classes: main text body, decorations and comments.
Finally, HBA 1.0~\cite{HBA} is one of the largest pixel-level segmentation datasets currently available. It includes 6 manuscripts and 5 printed books with segmentation into 6 classes according to different text fonts (e.g. lowercase, uppercase, italics). However, the GTs of this dataset are very noisy and therefore are inadequate to represent the real performance of the models used for the analysis of the corresponding documents.

In this paper we present U-DIADS-Bib a novel dataset for pixel-precise document layout analysis characterized by virtually noiseless GTs that distinguish between 6 different semantic regions, allowing us to overcome the previously mentioned limitation of currently available datasets.

\section{Description of U-DIADS-Bib dataset}\label{sec3}
U-DIADS-Bib is composed of 200 images, 50 for each of the 4 different manuscripts that characterize it. These handwritten books were selected in collaboration with humanist partners considering both the complexity of their layout and the presence of significant and semantically distinguishable elements. In particular, the images of the four manuscripts were collected from the French digital library Gallica\footnotemark[1]. All manuscripts are Latin and Syriac Bibles published between the 6th and 12th centuries A.D. which will be briefly described hereafter:
\begin{itemize}
\item Paris, Bibliothèque nationale de France, Latin 2\footnotemark[2]. The manuscript, known as the Second Bible of Charles the Bald, was produced between A.D. 871 and 877 at the Abbey of Saint-Amand (Haute-France); it was kept in the Abbey of Saint-Denis between A.D. 877 and 1595 and later transferred to the Royal Library of France.
It is composed of 444 parchment pages and the layout is structured in two columns.
\item Paris, Bibliothèque nationale de France, Latin 14396\footnotemark[3]. The manuscript was produced between A.D. 1145 and 1150 in the Abbey of Saint-Victor (Paris). It contains the biblical text from the Book of Ezra to the Book of Revelation and it is probably the final part of a three-volume Bible; this codex, known as Genesis-Kings, is the first volume.
The manuscript is composed of 170 parchment pages and the layout is structured in two columns.
\item Paris, Bibliothèque nationale de France, Latin 16746\footnotemark[4]. The manuscript was produced between A.D. 1170 and 1190 at the Abbey of Saint Bertin (Pas-de-Calais). It contains the New Testament and it is the final part of a four-volume Bible (Paris, Bibliothèque nationale de France Latin 16743-16746). The Bible had been kept for a long time in the Capuchin convent of Saint-Honoré, in Paris.
The manuscript is composed of 176 parchment pages and the layout is structured in two columns.
\item Paris, Bibliothèque nationale de France, Syriaque 341\footnotemark[5]. The manuscript was produced between the 6th and 7th centuries A.D. probably in the Monastery of Baquqa in Iraq. It contains the Old Testament in the Syriac Peshitta version and portions of the New Testament. The manuscript is composed of 256 parchment and paper pages, and the layout is structured in three columns.
\end{itemize}
\footnotetext[1]{Source https://gallica.bnf.fr}
\footnotetext[2]{https://gallica.bnf.fr/ark:/12148/btv1b8452767n}
\footnotetext[3]{https://gallica.bnf.fr/ark:/12148/btv1b84429190}
\footnotetext[4]{https://gallica.bnf.fr/ark:/12148/btv1b85144288}
\footnotetext[5]{https://gallica.bnf.fr/ark:/12148/btv1b10527102b}

For the dataset, we selected a representative set of 50 pages from each manuscript. A sample page for each manuscript and a corresponding detail view are shown in Fig.~\ref{fig:examplepage}.
\begin{figure}[t]%
    \centering
    \subfloat[\centering Main Text]{{\includegraphics[width=.42\linewidth]{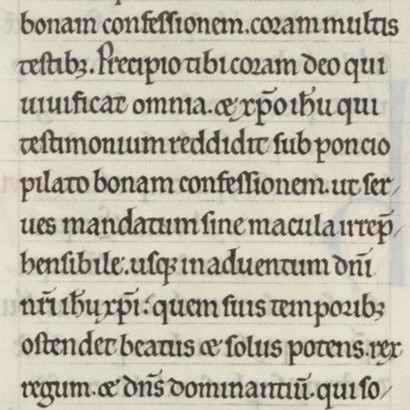} \label{fig:text} }}%
    \qquad
    \subfloat[\centering Decoration]{{\includegraphics[width=.42\linewidth]{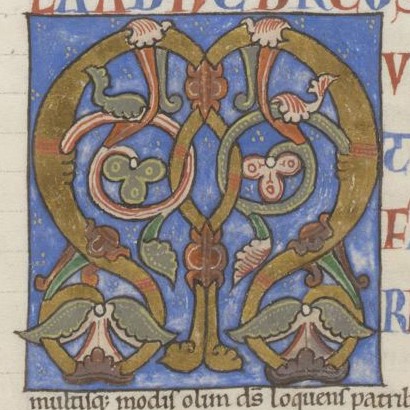}  \label{fig:decoration}}}%
    \qquad
    \subfloat[\centering Title]{{\includegraphics[width=.42\linewidth]{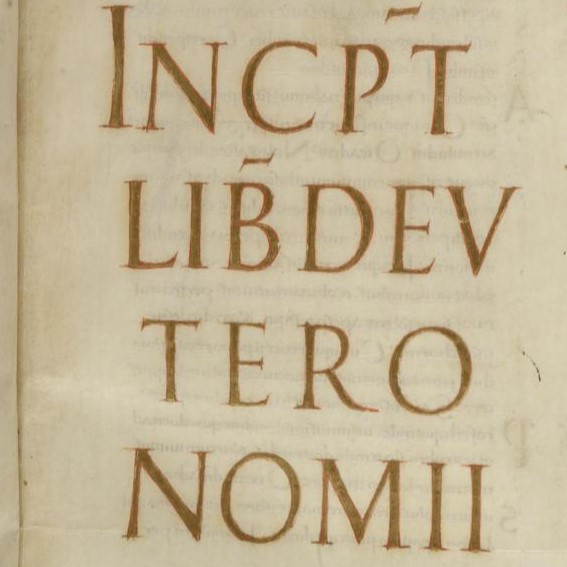} \label{fig:title}}}%
    \qquad
    \subfloat[\centering Chapter Headings]{{\includegraphics[width=.42\linewidth]{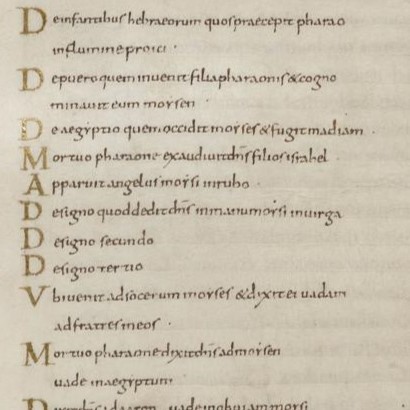} \label{fig:summary}}}
    \qquad
    \subfloat[\centering Paratext]{{\includegraphics[width=.42\linewidth]{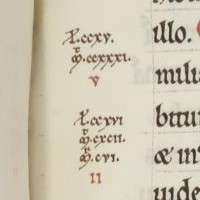} \label{fig:paratext}}}
    \qquad
    \subfloat[\centering Background]{{\includegraphics[width=.42\linewidth]{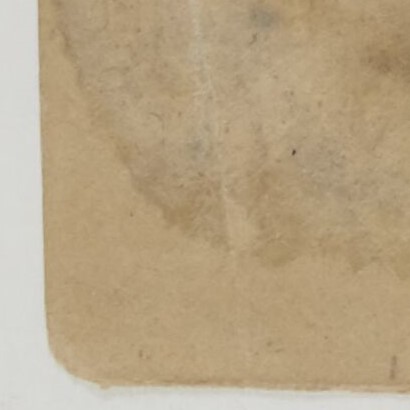} \label{fig:BG}}}%
    \caption{Samples of six segmentation classes: Main Text (\ref{fig:text}),   Decoration (\ref{fig:decoration}), Title (\ref{fig:title}), Chapter Headings (\ref{fig:summary}) , Paratext (\ref{fig:paratext}) and Background (\ref{fig:BG}).}%
    \label{fig:exampleclass}%
\end{figure}

\begin{figure*}[t]%
    \centering
    \subfloat[\centering Latin 2, page 144 original]{{\includegraphics[width=.205\linewidth]{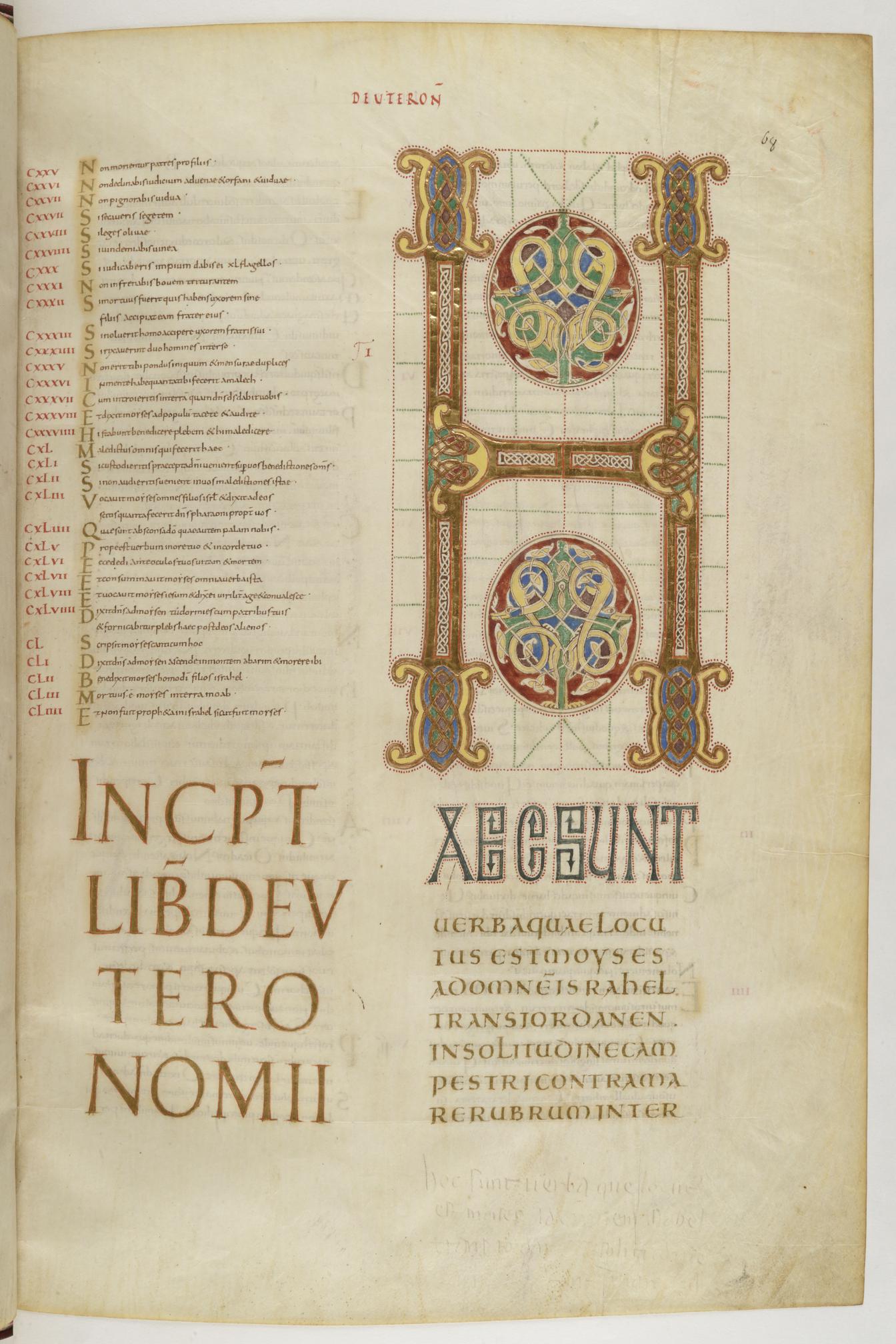} \label{fig:2orig} }}%
    \qquad
    \subfloat[\centering Latin 14396, page 325 original]{{\includegraphics[width=.205\linewidth]{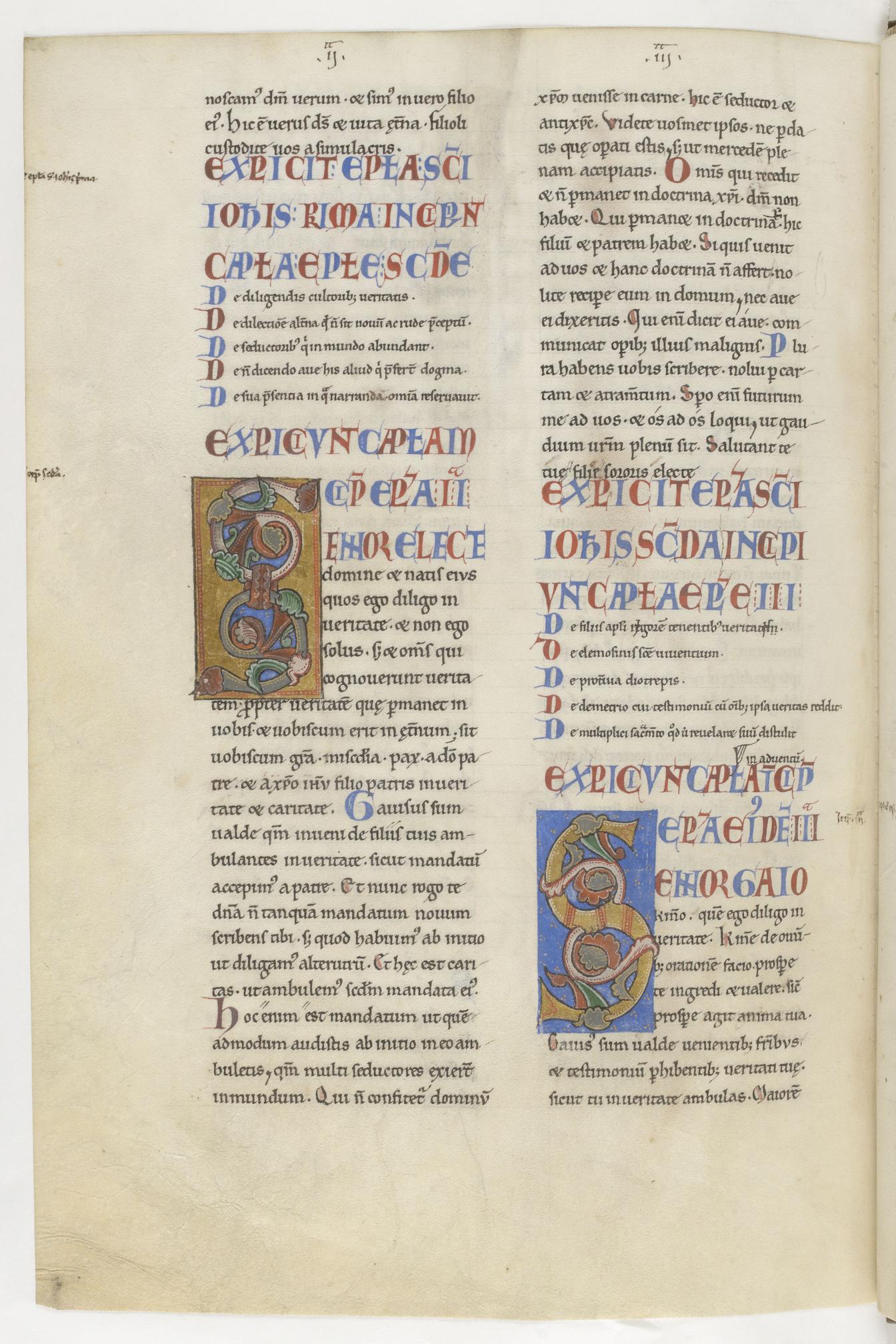} \label{fig:14396orig}}}
    \qquad
    \subfloat[\centering Latin 16746, page 187 original]{{\includegraphics[width=.205\linewidth]{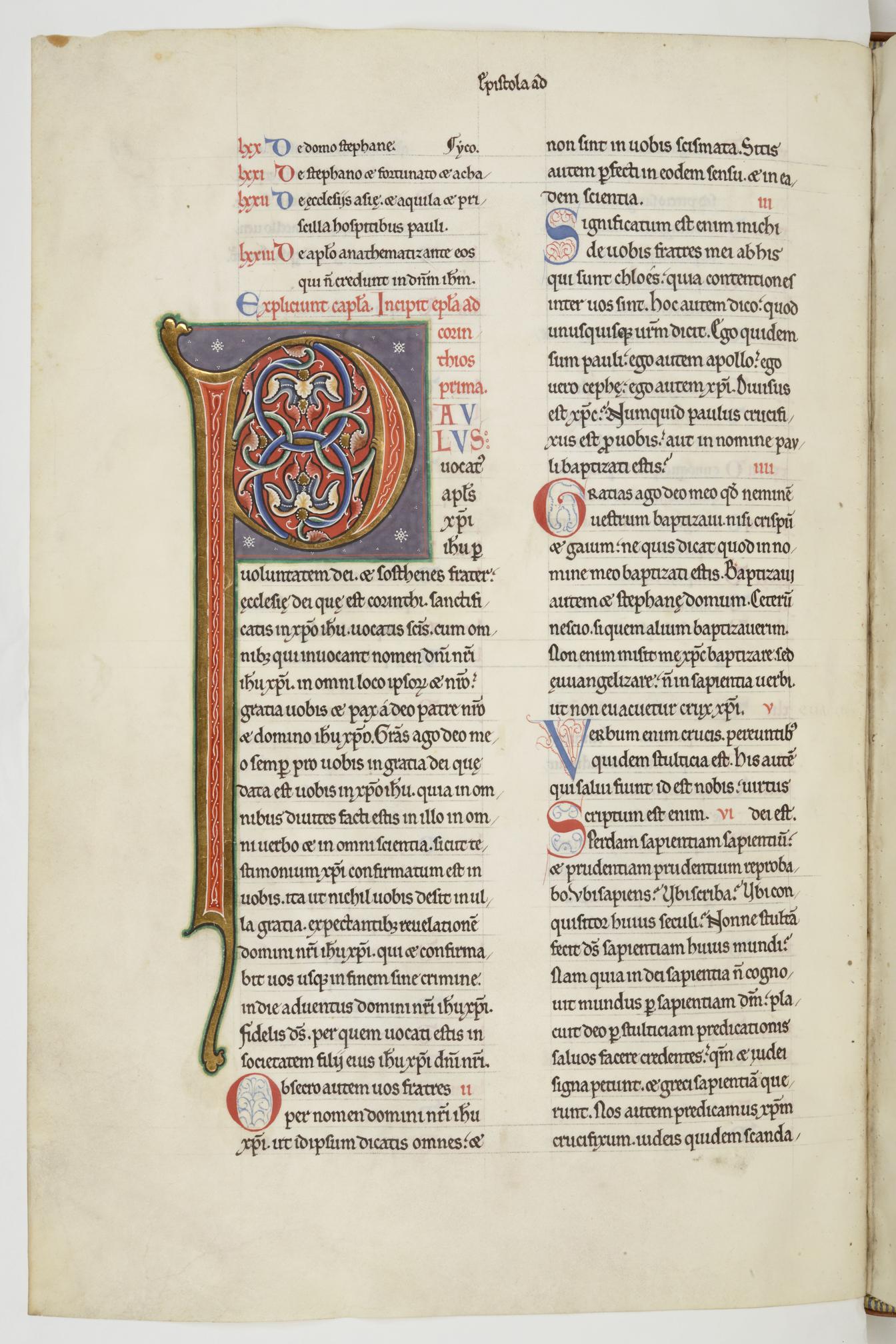}  \label{fig:16746orig}}}%
    \qquad
    \subfloat[\centering Syriaque 341, page 240 original]{{\includegraphics[width=.205\linewidth]{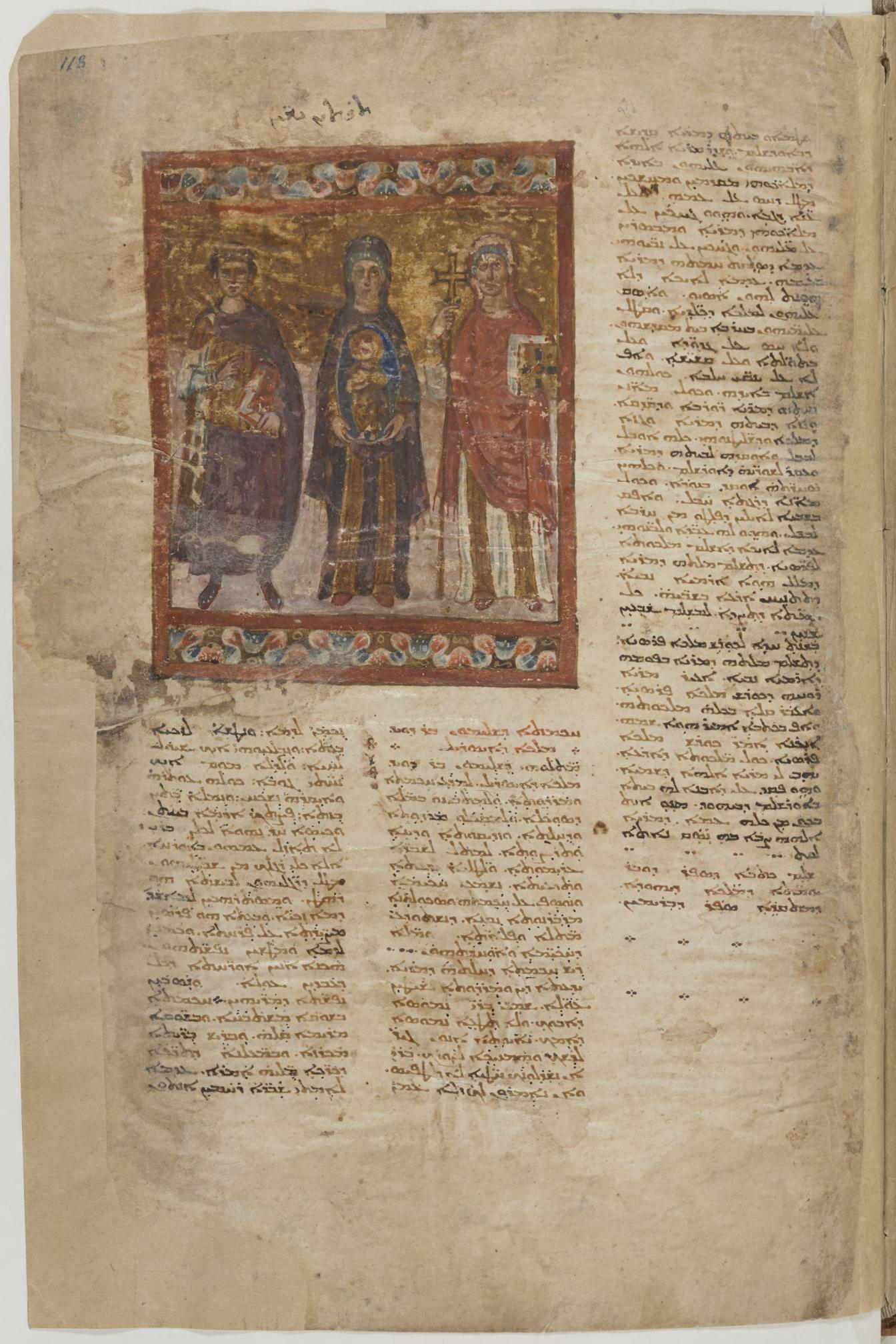}  \label{fig:341orig}}}%
    \qquad
    \subfloat[\centering Latin 2 ground truth]{{\includegraphics[width=.205\linewidth]{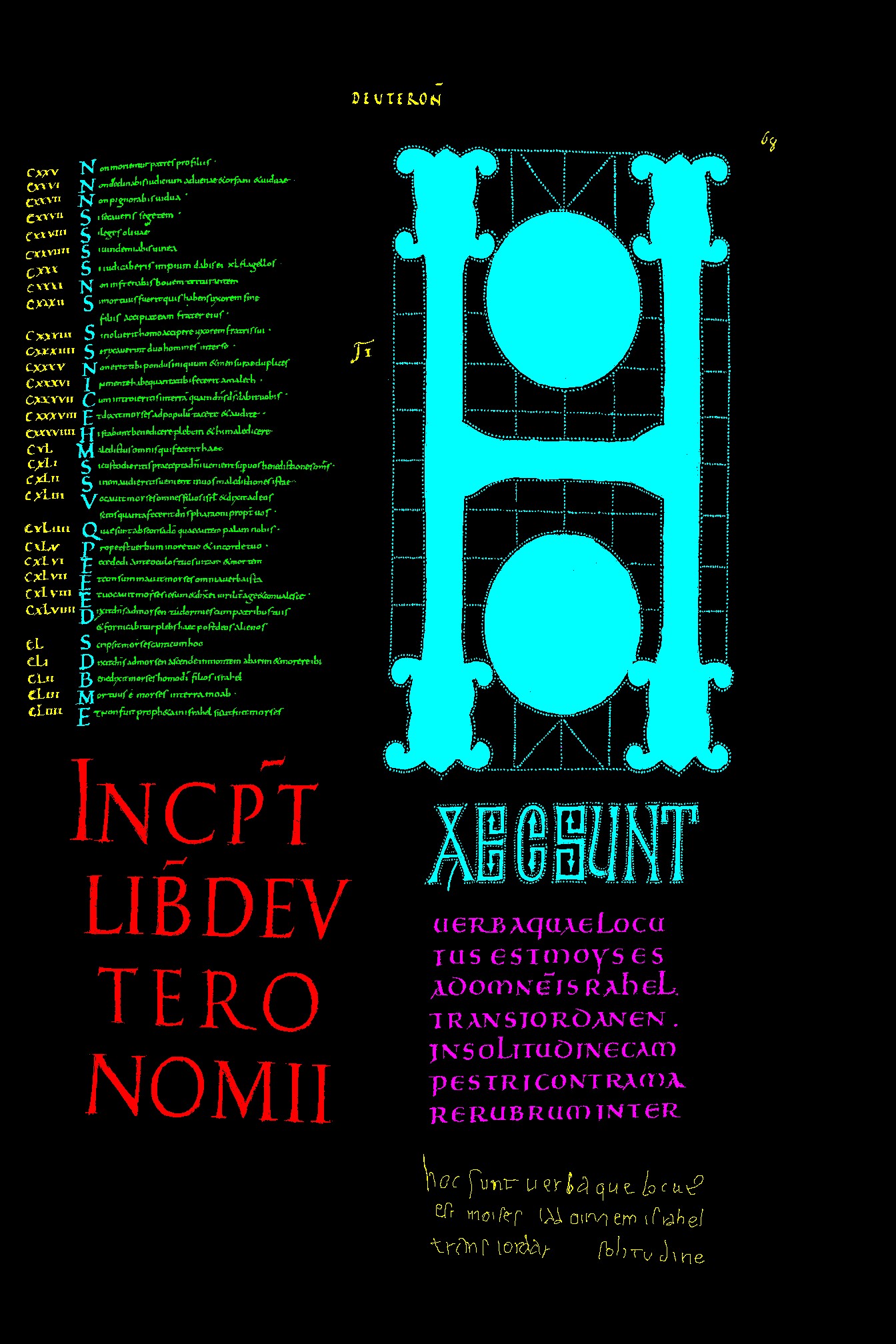} \label{fig:2GT}}}%
    \qquad
    \subfloat[\centering Latin 14396 ground truth]{{\includegraphics[width=.205\linewidth]{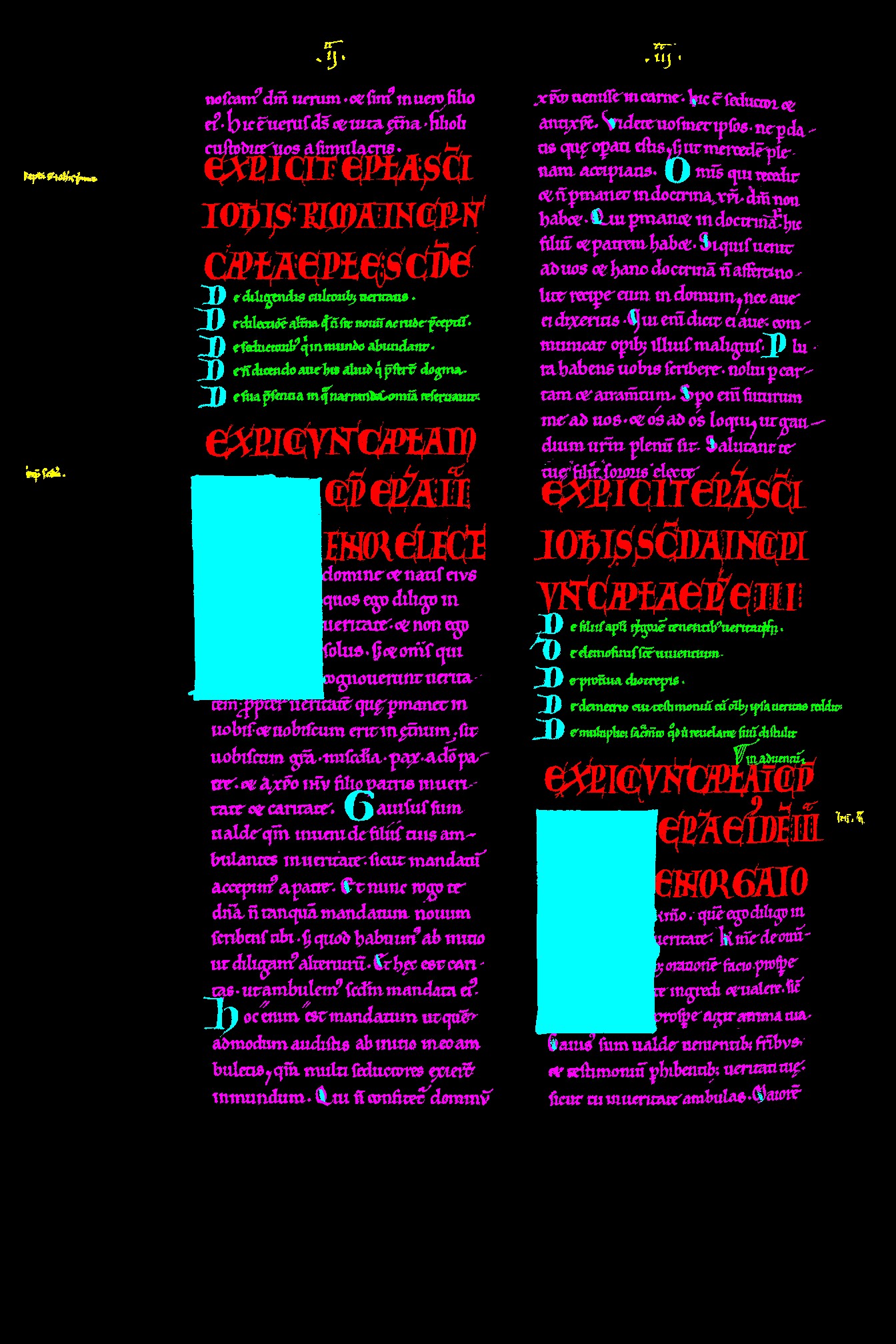} \label{fig:14396GT}}}
    \qquad
    \subfloat[\centering Latin 16746 ground truth]{{\includegraphics[width=.205\linewidth]{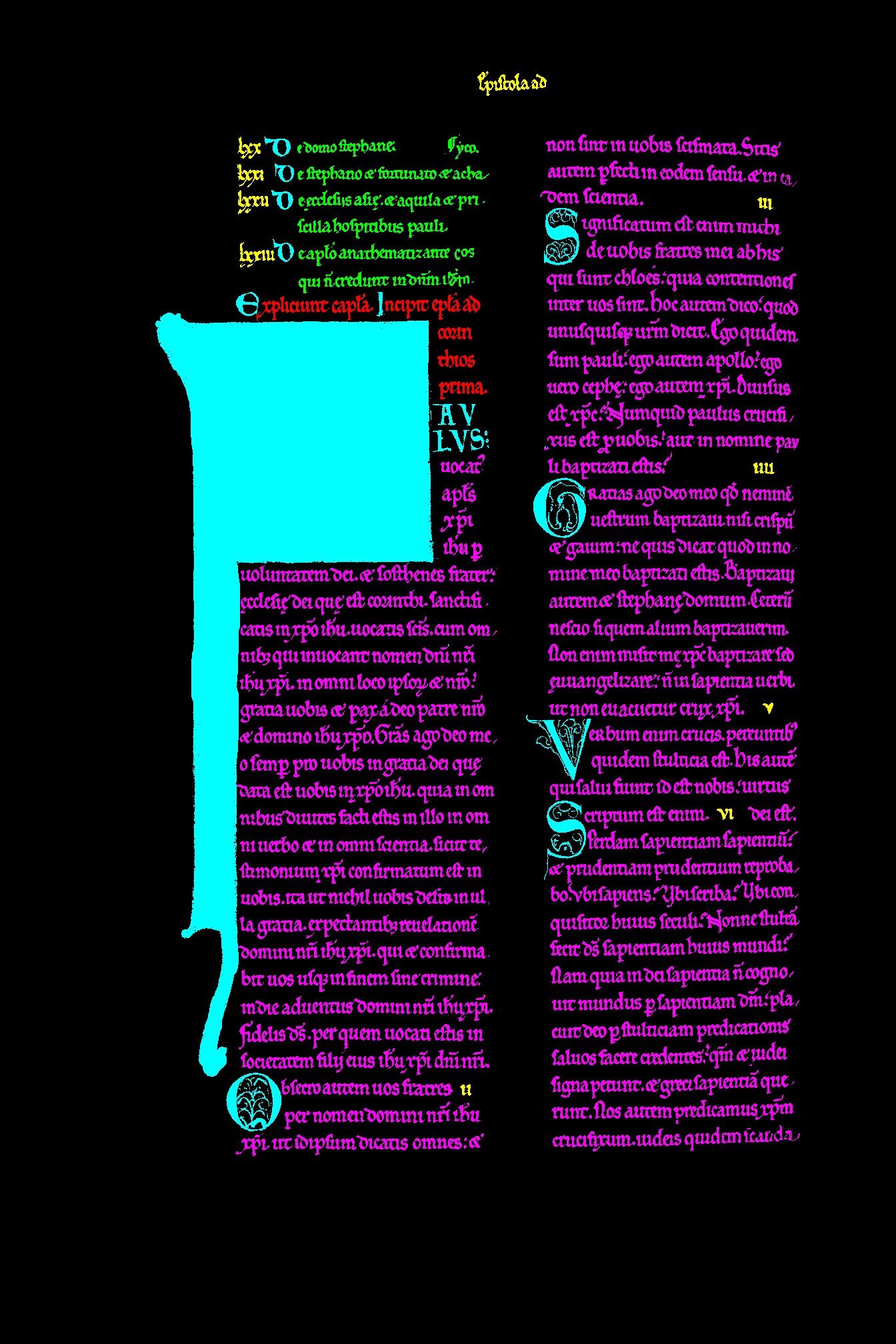} \label{fig:16746GT}}}
    \qquad
    \subfloat[\centering Syriaque 341 ground truth]{{\includegraphics[width=.205\linewidth]{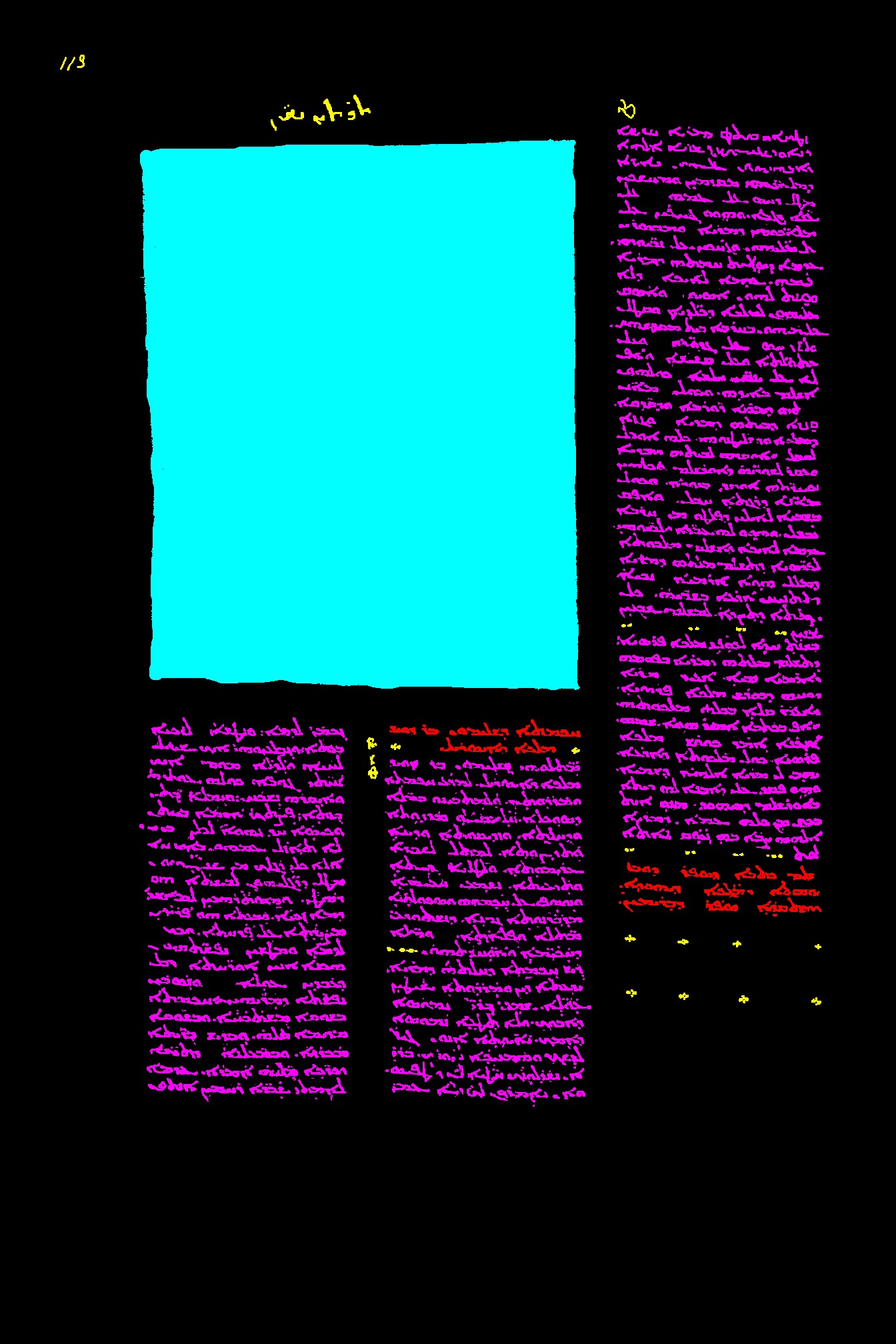} \label{fig:341GT}}}%
    \caption{Images showing a page of each manuscript (~\ref{fig:2orig}\textendash~\ref{fig:341orig}) as well as its corresponding ground truth mask (\ref{fig:2GT}\textendash~\ref{fig:341GT}), in which each color represents a different semantic class of the document layout.}%
    \label{fig:exampleGT}%
\end{figure*}

\subsection{Segmentation Analysis}
The choice of biblical texts with the homogeneity of the contents, and the abundance of material from different centuries, scriptures and layouts that characterize them, allows for the challenge of better highlighting the differences in the graphic and textual elements that accompany the text but are not part of it.
It is important to remark that these decisions were supported by scholars in the humanities, which are experts in the respective area.
Indeed, from a historical perspective, the Bible holds immense cultural significance as one of the most widespread and influential texts in the world. It has profoundly impacted human societies across different spatial and temporal contexts. In particular, the biblical manuscripts represent a rich diversity of historical layouts and textual variations, due to their wide diffusion over the centuries and in the world.

The six segmentation classes highlighted by humanist experts for page segmentation task are visible in Fig.\ref{fig:exampleclass} and are:
\begin{itemize}
\item Main Text. This class includes the writing area and represents the core and central content of the book. This class includes punctuation and pause marks. It can be structured in different layouts such as one or two columns.
\item Decoration. This class includes both figurative elements in the proper sense (miniatures and decorated initials), and elements of minimal decoration, such as initials with a graphic element or color distinguishing them from the rest of the text.
\item Title. More properly \textit{'incipit and explicit formulae'}, identifiable by the use of a different ink color and/or by the adoption of display scripts: monumental (square) or rustic capital, uncials or mixed capital/uncials script. 
\item Chapter Headings. The chapter headings function in ancient manuscripts was to facilitate the retrieval of a particular chapter or passage, but according to different scansion and interpretation from one set to another. From a graphical point of view, they are often recognizable by a script similar to that of the main text but smaller in size.
\item Paratext. This class consists of several elements such as glosses, marginal and interlinear notes, corrections, paragraph numbering, possession notes from different periods, page or fascicle numbering. In general, includes all hand annotations that do not fit into the other classes.
\item Background. This class includes the background of the page and any outline visible in the scanned image.
\end{itemize}

However, there is also a twofold motivation behind this choice from a humanistic point of view: firstly, the classification of paratexts is in many cases neither unambiguous nor immediate (e.g. it is not easy to distinguish at a glance a textual variant from the integration of a passage omitted by mistake in the original text or the correction of an erroneous text), and the use of broader classes avoids stiffening the interpretation of paratexts into dead ends. Furthermore, the broader aim is to encourage a wider use of document layout analysis from an interdisciplinary point of view: the extraction of paratextual elements of different natures can in fact provide material for scholars of different historical, textual and librarian disciplines, from philology to paleography to art history to the history of ancient and modern libraries.

\subsection{Ground Truth construction}

The annotation of the U-DIADS-Bib is the result of the collaboration between humanists and computer scientists. As previously addressed, manual annotation of images is a very time-consuming task, especially in the context of document segmentation where the different layout components can be very small and detailed, whilst annotations provided by algorithms tend to present many inaccuracies and are prone to the introduction of noise.
Consequently, since our goal was to produce a dataset with a large number of annotated pages, pixel-precise segmentation and very limited, if any, noise, we have defined a segmentation pipeline that involves the alternation of humanist and algorithmic work, in order to optimize the expected results.

First of all, after having chosen the manuscripts, a subset of 50 images was selected in such a way as to represent all the chosen segmentation classes for each manuscript.
A subset of 10 images per manuscript was selected and binarized using Sauvola threshold technique~\cite{SAUVOLA} and morphological operators in order to give the human experts a starting point to work on. 
Then, experts in manuscript texts have manually segmented at pixel-level with different colors these few images per manuscript contain examples of all expected classes.
The next step was to train a machine learning model with these subsets of images and the related GTs segmented by humanists to obtain a coarse segmentation of the entire dataset.
To achieve this, the framework proposed in~\cite{de2023few} was followed, where a few-shot pixel-precise document layout segmentation method with high performance is presented.
The segmentation was done for 4 classes, so as to obtain a less detailed but well-defined segmentation on the whole dataset and with the presence of almost zero noise. Finally, the expert humanists introduced the other missing semantic classes and meticulously refined and corrected all the color masks of the GTs by comparing them to the original images. It is worth noting that, despite the task being computer-aided, the final result is always defined by a human expert, thus avoiding possible biases or errors in the dataset.

Fig.~\ref{fig:exampleGT} illustrate some examples of the defined GT and corresponding original image for each manuscript of the U-DIADS-Bib. Each selected pixel is marked by a color that symbolizes the corresponding content type.

U-DIADS-Bib is therefore composed of 50 original color page images for each manuscript, stored in JPEG image format with resolution $1344\times2016$ px.
Each page is associated with the corresponding GT data, stored in a PNG image with the same size of the original one.
GTs contain six different and non-overlapping annotated classes (background (BG), comment, decoration, text, title and chapter headings (CH)) encoded by RGB value as follow:
\begin{itemize}
\item RGB(0,0,0) Black: Background
\item RGB(255,255,0) Yellow: Paratext
\item RGB(0,255,255) Cyan: Decoration
\item RGB(255,0,255) Magenta: Main Text
\item RGB(255,0,0) Red: Title
\item RGB(0,255,0) Lime: Chapter Headings
\end{itemize}

The Syriaque 341 manuscript contains five semantic classes and the missing one is the Chapter Headings.

For each of the manuscripts included U-DIADS-Bib, 10 images have been selected for the training set, 10 for the validation set and the remaining 30 for the test.
As shown in Table~\ref{tab:classdistr} the proposed dataset is characterized by a noticeable imbalance between the different layout classes, making the layout segmentation task even more challenging, due to the fact that the classes that appear less consistently in the layout of the selected pages are much harder to identify for the models employed
for the segmentation task
\begin{table}[htb]
\centering
\setlength{\tabcolsep}{3pt}
\begin{tabular}{ccccc}
\hline & \textbf{Lat 2} & \textbf{Lat 14396} & \textbf{Lat 16746}& \textbf{Syr 341}\\\hline
\textbf{BG} & 92.8 &89.2 & 88&85.1\\
\textbf{Paratext} &0.1 &0.1 & 0.3&0.2 \\
\textbf{Decoration} & 1.5& 2&3&2.8\\
\textbf{Main Text}& 4.7 &  7.6&7.8 &11.9\\
\textbf{Title}& 0.4& 0.5&0.1&0.1\\
\textbf{CH} &0.5&0.6&0.8&0\\
\hline
\textbf{Total} &100&100&100&100\\
\hline
\end{tabular}
\caption{Classes distribution (\%) at pixel level for each manuscripts of U-DIADS-Bib.}
\label{tab:classdistr}
\end{table}

Moreover, as we believe that real-world use of layout analysis techniques will be broadly adopted only if the amount of required manually-labeled training data is reduced, we also provided a standardized few-shot version of our dataset (U-DIADS-BibFS), which we hope will encourage further efforts towards this goal.
U-DIADS-BibFS offers 43 images per manuscript, where only 3 images are provided for training, 10 for validation and 30 for testing.
For each manuscript, the 3 training images were appropriately chosen so that they included all the segmentation classes.
The detail of the semantic component distributions of this few-shot dataset are shown in Table~\ref{tab:classdistrFS}.

\begin{table}[htb]
\centering
\setlength{\tabcolsep}{3pt}
\begin{tabular}{ccccc}
\hline & \textbf{Lat 2} & \textbf{Lat 14396} & \textbf{Lat 16746}& \textbf{Syr 341}\\\hline
\textbf{BG} & 92.9 & 89.2& 88&85.1\\
\textbf{Paratext} &0.1 & 0.1& 0.3 &0.2\\
\textbf{Decoration} & 1.3& 2&3&2.6\\
\textbf{Main Text}& 4.8 &  7.5&7.8 &12\\
\textbf{Title}& 0.4& 0.5&0.1&0.1\\
\textbf{CH} &0.5&0.7&0.8&0\\
\hline
\textbf{Total} &100&100&100&100\\
\hline
\end{tabular}
\caption{Classes distribution (\%) at pixel level for each manuscripts of U-DIADS-BibFS.}
\label{tab:classdistrFS}
\end{table}
The dataset is freely available for download on the U-DIADS-Bib repository\footnotemark[6].

\footnotetext[6]{https://ai4ch.uniud.it/udiadsbib/}

\begin{table*}[htb]
\Huge
\centering
\resizebox{\textwidth}{!}{
\begin{tabular}{lcccccc}
\hline
Manuscript& \multicolumn{1}{c}{\begin{tabular}[c]{@{}c@{}}Metric\\ (weight/macro)\end{tabular}}  & \multicolumn{5}{c}{Backbone}\\\hline
                          &  & FCN~\cite{fcn} & LRSAPP~\cite{lraspp} & DeepLabV3~\cite{deeplab} & DeepLabV3+~\cite{deeplabv3plus} & PSPNET~\cite{pspnet} \\\hline
\multirow{4}{*}{Latin 2}  & Precision & 0.955/0.542& 0.953/0.489& 0.954/0.491 &\textbf{0.962}/\textbf{0.593}& 0.932/0.217\\
                          & Recall & 0.926/0.708& 0.898/0.499 & 0.907/0.811& \textbf{0.936}/\textbf{0.828}& 0.890/0.297  \\
                          & IoU & 0.892/0.425& 0.862/0.373& 0.870/0.451& \textbf{0.906}/\textbf{0.544}& 0.851/0.200 \\
                          & F1-Score & 0.935/0.535& 0.916/0.460& 0.923/0.585& \textbf{0.945}/\textbf{0.670}& 0.904/0.235  \\\hline

\multirow{4}{*}{Latin 14396} & Precision & 0.953/0.519& 0.939/0.616 &0.951/0.590 &\textbf{0.961}/\textbf{0.659} &0.951/0.560 \\
                          & Recall &   0.936/0.752& 0.906/0.589& 0.918/\textbf{0.902}& \textbf{0.944}/0.842& 0.917/0.779 \\
                          & IoU &   0.898/0.500& 0.867/0.417& 0.875/0.556& \textbf{0.910}/\textbf{0.609}& 0.873/0.531 \\
                          & F1-Score &  0.940/0.586& 0.919/0.531& 0.928/0.683& \textbf{0.949}/\textbf{0.730}& 0.927/0.633 \\\hline

\multirow{4}{*}{Latin 16746} & Precision &  0.948/0.564& 0.945/0.614& 0.947/0.586& \textbf{0.958}/\textbf{0.664}& 0.926/0.389 \\
                          & Recall &  0.937/0.787& 0.893/0.892& 0.905/\textbf{0.953}& \textbf{0.939}/0.932& 0.899/0.460 \\
                          & IoU &  0.897/0.541& 0.842/0.555& 0.856/0.560& \textbf{0.900}/\textbf{0.633}& 0.845/0.362 \\
                          & F1-Score &  0.939/0.638& 0.908/0.694& 0.917/0.693& \textbf{0.944}/\textbf{0.757}& 0.906/0.412 \\\hline

\multirow{4}{*}{Syriaque 341} & Precision &  0.942/0.669& 0.912/0.575& 0.927/0.625& \textbf{0.943}/\textbf{0.745}& 0.916/0.491\\
                              & Recall &   0.928/\textbf{0.851}& 0.878/0.436& 0.905/0.739& \textbf{0.931}/0.686& 0.899/0.528\\
                              & IoU &  0.879/\textbf{0.600}& 0.808/0.353& 0.846/0.551& \textbf{0.883}/0.530& 0.837/0.448 \\
                              & F1-Score &  0.932/\textbf{0.719}& 0.885/0.433& 0.912/0.665& \textbf{0.934}/0.617& 0.904/0.509 \\\hline

\multirow{4}{*}{Mean} & Precision & 0.950/0.574& 0.937/0.574& 0.945/0.573& \textbf{0.956}/\textbf{0.665}& 0.931/0.414 \\
                      & Recall  &  0.932/0.775& 0.894/0.604& 0.909/\textbf{0.851}& \textbf{0.938}/0.822& 0.901/0.516    \\
                      & IoU  &  0.892/0.517& 0.845/0.425& 0.862/0.530& \textbf{0.900}/\textbf{0.579}& 0.852/0.385  \\
                      & F1-Score   &  0.937/0.620& 0.907/0.530& 0.920/0.657& \textbf{0.943}/\textbf{0.694}& 0.910/0.447 \\\hline

\end{tabular}}
\caption{Comparison of the weighted average and macro average performance of the different semantic segmentation models chosen tested on U-DIADS-Bib dataset. The best-performing results for each manuscript and for full dataset are shown in bold.}
\label{tab:wholedatasetresult}
\end{table*}

\section{Benchmarking setup}\label{sec4}
In order to analyze our data and to provide a benchmark for future studies, we selected a set of 5 popular deep-learning-based semantic segmentation approaches, namely FCN~\cite{fcn}, Lite Reduced Atrous Spatial Pyramid Pooling (LRASPP)~\cite{lraspp}, DeepLabV3~\cite{deeplab}, its improvement represented by DeepLabV3+~\cite{deeplabv3plus}, and Pyramid Scene Parsing Network (PSPNet)~\cite{pspnet}.
Also, we include the results obtained by state-of-the-art for the task of document layout segmentation in a few-shot learning setting, presented by De Nardin et al.~\cite{WACV}.

\subsection{Training Setup}
All the models were trained using the same setup: the Adam optimizer with a learning rate of $10^{-3}$ and a weight decay of $10^{-5}$.
The models are trained for a maximum of 200 epochs with an early stop mechanism that monitors the validation loss at every epoch and stops the training process if it did not improve over the last 20 consecutive epochs. However, to ensure that the model is trained for a minimum number of iterations, thus reducing the risk of convergence to a local minimum at the beginning of the training process, a buffer of 50 epochs is added.
Due to the high imbalance between the semantic classes in each document category of the dataset, we have decided to adopt a weighted cross-entropy loss function (Eq.~\ref{eq:weights}), where the weight for each class is inversely proportional to the frequency of that element in the corresponding manuscript ($F_i$). The idea behind this choice is to allow the selected segmentation models to focus more on the classes that are less represented in the dataset, therefore mitigating the challenge introduced by its imbalanced nature.
\begin{equation}
    W_i = \sqrt{\frac{1}{F_i}}
    \label{eq:weights}
\end{equation}
All models rely on a residual network as their backbone, in particular the ResNet34 version, except LRASPP which is based on MobileNetV3-Large architecture.

\begin{figure*}[tb]%
    \centering
    \begin{adjustbox}{minipage=\linewidth,scale=0.8}
    \subfloat[\centering Latin 2]{{\includegraphics[width=.45\linewidth]{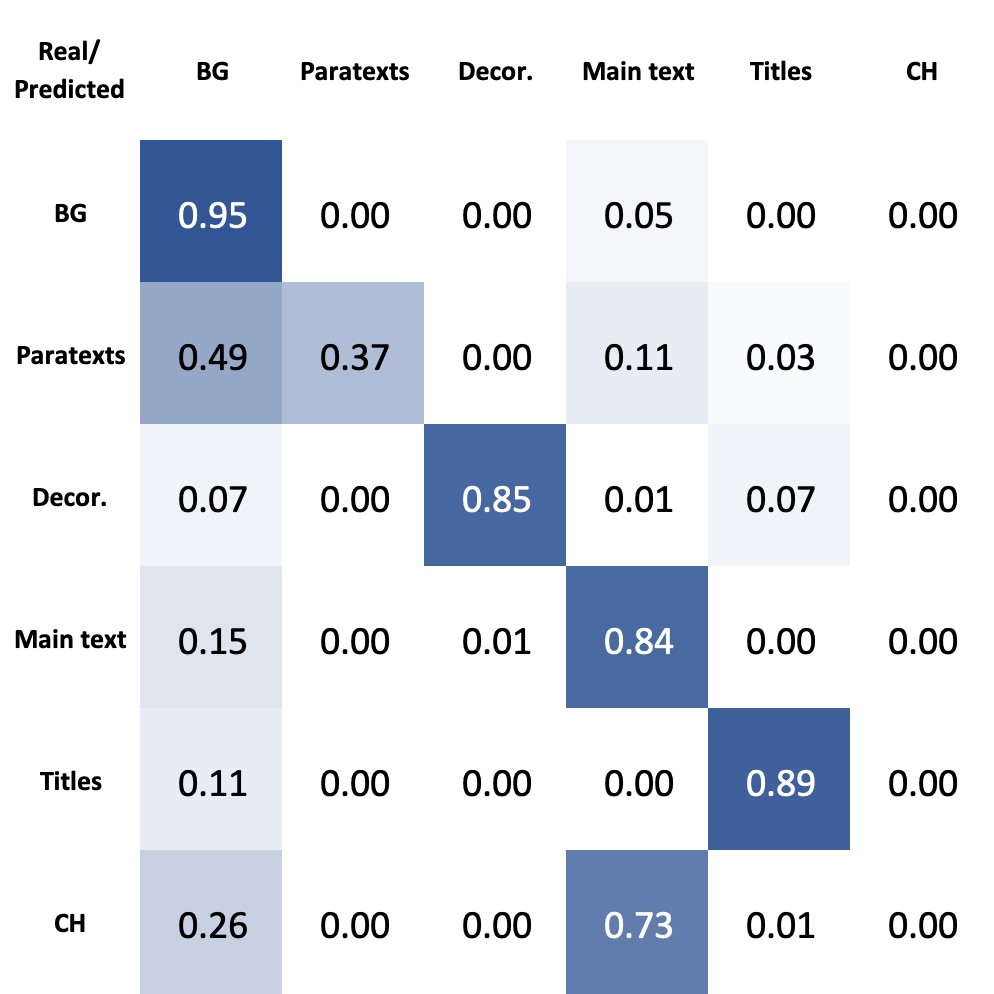} \label{fig:l2conf} }}%
    \subfloat[\centering Latin 14396]{{\includegraphics[width=.45\linewidth]{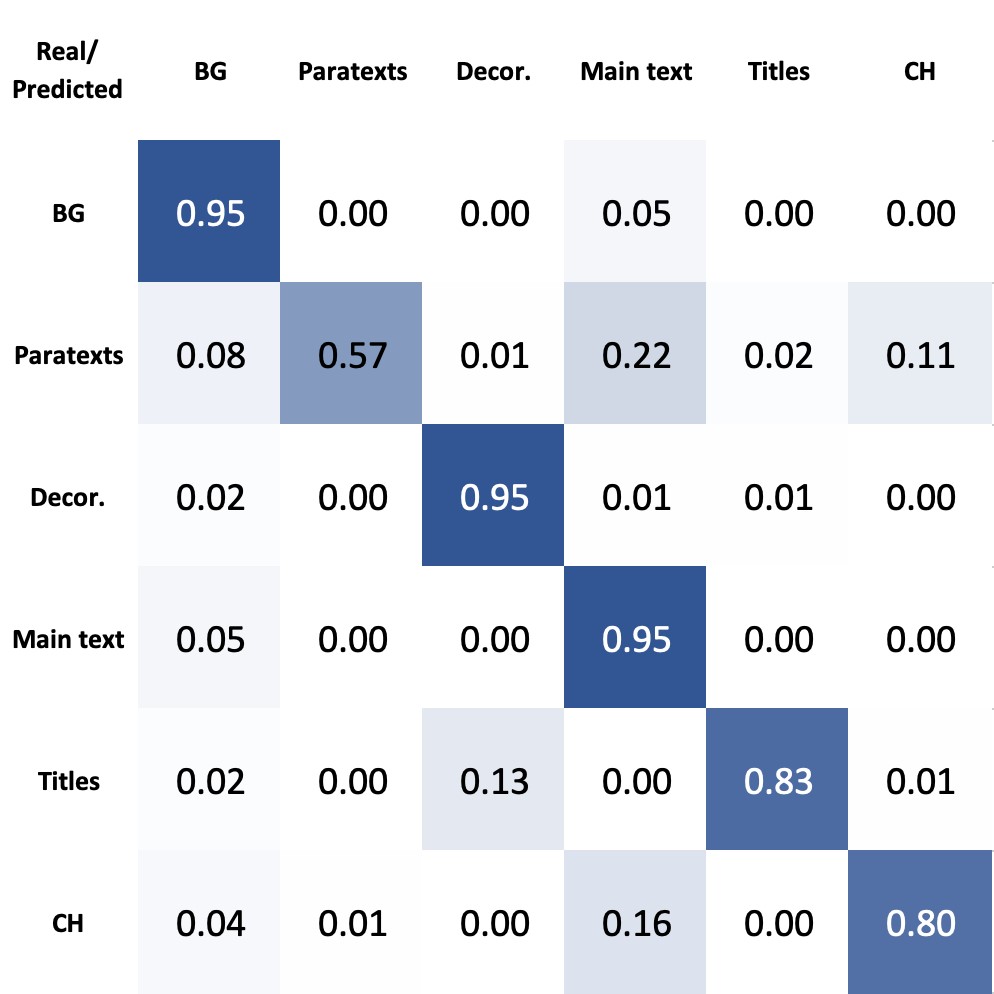} \label{fig:l14conf}}}
    \\
    \subfloat[\centering Latin 16746]{{\includegraphics[width=.45\linewidth]{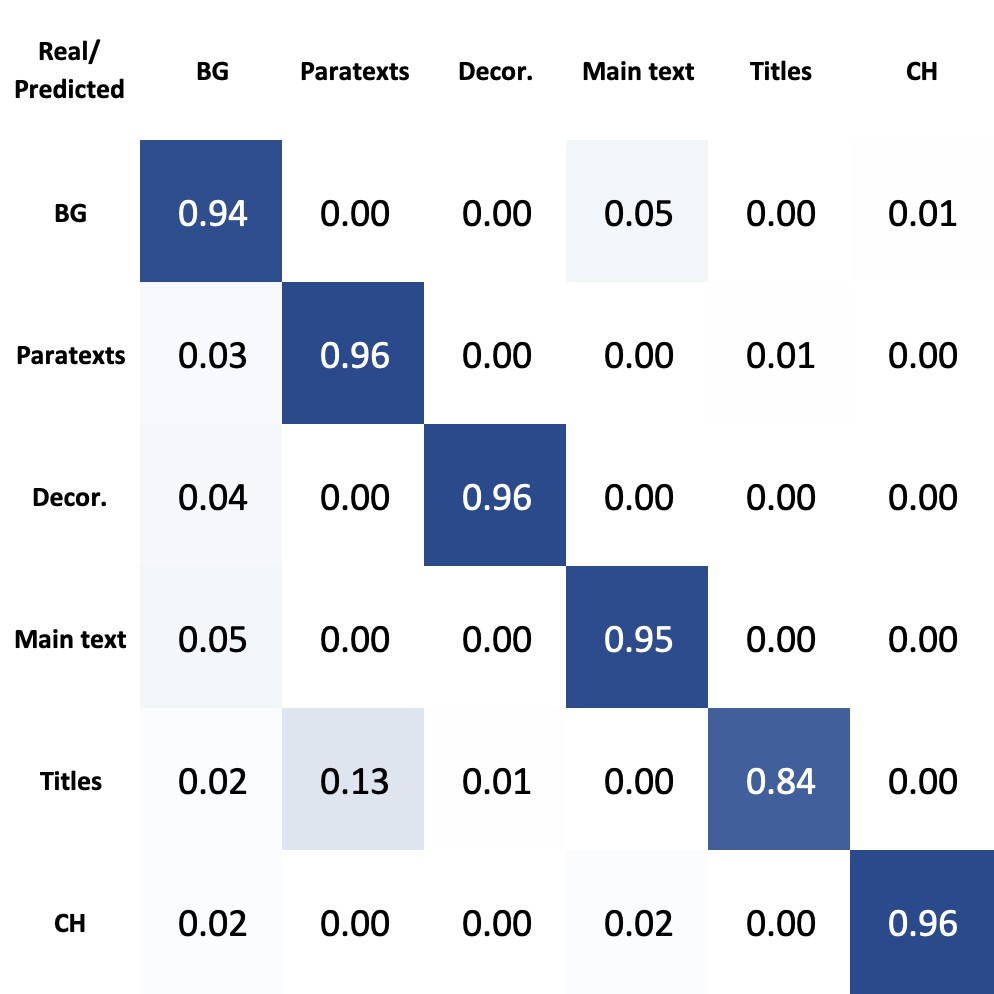} \label{fig:l16conf} }}%
    \subfloat[\centering Syriaque 341]{{\includegraphics[width=.45\linewidth]{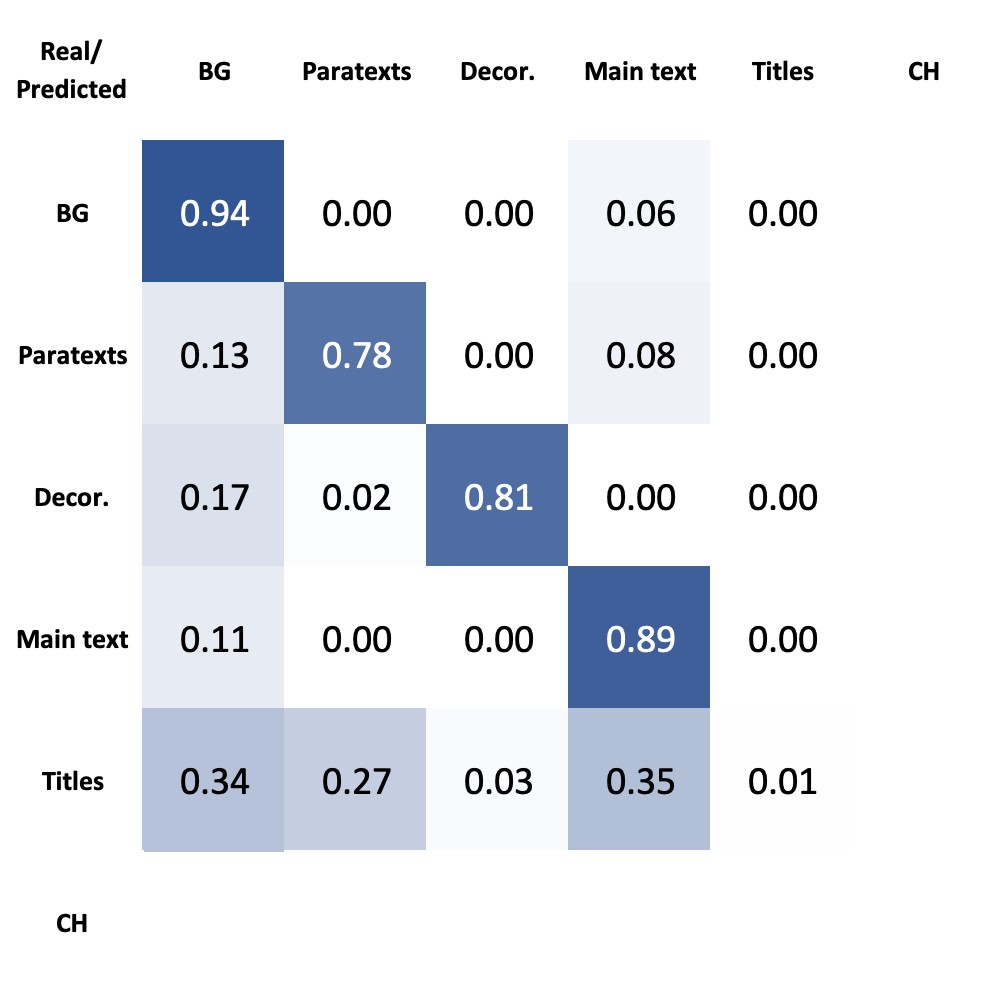} }}
    \caption{Confusion matrices obtained by applying  DeepLabV3+\cite{deeplabv3plus} to the four ancient manuscripts of the U-DIADS-Bib dataset.}
    \label{fig:confmatrix}%
    \end{adjustbox}
\end{figure*}

\begin{table*}[htb]
\Huge
\centering
\resizebox{\textwidth}{!}{
\begin{tabular}{lccccccc}
\hline
\multicolumn{1}{c}{\begin{tabular}[c]{@{}c@{}}Manuscript\\ Few-Shot\end{tabular}} & \multicolumn{1}{c}{\begin{tabular}[c]{@{}c@{}}Metric\\ (weight/macro)\end{tabular}}  & \multicolumn{6}{c}{Backbone}\\\hline
                          && FCN~\cite{fcn} & LRSAPP~\cite{lraspp} & DeepLabV3~\cite{deeplab} & DeepLabV3+~\cite{deeplabv3plus} & PSPNET~\cite{pspnet} & De Nardin et al.~\cite{WACV}\\\hline
\multirow{4}{*}{Latin 2}  & Precision & 0.873/0.225 & 0.907/0.559 &0.939/0.442 &0.958/0.537 &0.928/0.355 & \textbf{0.973}/\textbf{0.636} \\
                          & Recall & 0.930/0.194 &0.927/0.203 &0.897/0.570 &0.932/0.756 &0.868/0.296 &\textbf{0.972}/\textbf{0.787} \\
                          & IoU & 0.867/0.177 &0.909/0.186 &0.863/0.369 &0.902/0.485 &0.887/0.190 &\textbf{0.950}/\textbf{0.547}\\
                          & F1-Score & 0.899/0.200 &0.871/0.216 &0.914/0.467 &0.941/0.602 &0.822/0.227 &\textbf{0.970}/\textbf{0.685} \\\hline

\multirow{4}{*}{Latin 14396} & Precision & 0.845/0.426  &   0.852/0.398  &  0.945/0.549    &  0.958/0.705    &    0.925/0.377 & \textbf{0.971}/\textbf{0.719}   \\
                          & Recall &  0.908/0.612  &    0.888/0.234    &  0.908/\textbf{0.794}    &  0.941/0.789     &  0.912/0.447 & \textbf{0.973}/0.718     \\
                          & IoU &  0.833/0.408   &   0.866/0.206     & 0.863/0.496     &  0.905/0.612     & 0.915/0.350 &
                          \textbf{0.949}/\textbf{0.629}\\
                          & F1-Score &   0.874/0.485  &    0.808/0.257    &  0.920/0.618    &  0.946/\textbf{0.736} &  0.860/0.405  & \textbf{0.970}/0.707    \\\hline

\multirow{4}{*}{Latin 16746} & Precision &  0.839/0.312   &   0.897/0.472     &  0.934/0.525    &   0.941/0.560    &  0.875/0.246  & \textbf{0.952}/\textbf{0.714}    \\
                          & Recall &  0.880/0.553   &    0.869/0.547    &  0.887/\textbf{0.895}    &   0.911/0.731    &   0.858/0.467 & \textbf{0.959}/0.743    \\
                          & IoU &  0.810/0.289   &   0.867/0.288     &  0.839/0.483    &  0.863/0.464     & 0.857/0.228 & \textbf{0.921}/\textbf{0.636}      \\
                          & F1-Score &  0.858/0.341   &    0.796/0.378    &   0.904/0.610   &  0.920/0.598     &   0.789/0.282 & \textbf{0.951}/\textbf{0.761}     \\\hline

\multirow{4}{*}{Syriaque 341} & Precision & 0.755/0.320
&0.861/0.308 &0.890/0.487 &0.937/\textbf{0.654} &0.895/0.476 &\textbf{0.946}/0.602  \\
                              & Recall & 0.862/0.329
&0.883/0.321 &0.864/0.502 &0.926/0.520 &0.862/0.436&\textbf{0.949}/\textbf{0.549}  \\
                              & IoU &  0.745/0.277
&0.802/0.269 &0.789/0.365 &0.873/0.456 &0.787/0.354 &\textbf{0.907}/\textbf{0.473}  \\
                              & F1-Score &  0.804/0.323
&0.872/0.314 &0.872/0.463 &0.928/0.546 &0.871/0.436 &\textbf{0.946}/\textbf{0.564}   \\\hline

\multirow{4}{*}{Mean} & Precision &  0.828/0.321
&0.879/0.434 &0.927/0.501 &0.949/0.614 &0.906/0.364 &\textbf{0.961}/\textbf{0.668}  \\
                   & Recall  &  0.895/0.422
&0.917/0.326 &0.889/0.690 &0.928/\textbf{0.699} &0.875/0.412 &\textbf{0.963}/\textbf{0.699}  \\
                      & IoU  &   0.829/0.288
&0.879/0.237 &0.859/0.428 &0.900/0.504 &0.883/0.281 &\textbf{0.942}/\textbf{0.571}  \\
                      & F1-Score   &  0.844/0.337
&0.819/0.291 &0.882/0.540 &0.920/0.621 &0.815/0.338 &\textbf{0.950}/\textbf{0.679}  \\\hline

\end{tabular}}
\caption{Comparison of the weighted average and macro average performance of the different semantic segmentation models chosen and the current state-of-the-art tested on U-DIADS-BibFS dataset. The best-performing results for each manuscripts and for full dataset are shown in bold.}
\label{tab:fewshotresult}
\end{table*}

\subsection{Evaluation Metric}
The performances for these semantic segmentation tasks are measured using Precision, Recall, Intersection over Union (IoU) and F1-Score.
Metric definitions are reported in Eq.~\ref{prec}\textendash~\ref{f1}, where TP, FP and FN stand respectively for True Positives, False Positives and False Negatives. 
\begin{align}
    &\text{Precision} = \frac{\text{TP}}{\text{TP}+\text{FP}} \label{prec}\\
    &\text{Recall} = \frac{\text{TP}}{\text{TP}+\text{FN}} \label{rec}\\
    &\text{IoU} = \frac{\text{TP}}{\text{TP}+\text{FP}+\text{FN}} \label{iou}\\
    &\text{F1-Score} = \frac{2 \times \text{Precision} \times \text{Recall}}{\text{Precision} + \text{Recall}} \label{f1}
\end{align}

For each metric a weighted average is performed, based on each class distribution in each manuscript. Furthermore, the macro average for each metric is also presented to consider the weight of all segmentation classes equally important.
These evaluation metrics are calculated individually for each manuscript and a final evaluation of a model is obtained by averaging the metrics of the four handwritten documents.

\subsection{Results}
Table~\ref{tab:wholedatasetresult} shows the different performances of the aforementioned semantic segmentation models on the proposed U-DIADS-Bib dataset.
The results are provided both for a single manuscript and for the full dataset by averaging the scores obtained for each manuscript class.
In particular, both the values of the weighted average and the macro average of the metrics are presented. The reason behind this decision is that a consensus on which of the two averaging approaches is better suited for the layout segmentation task has not been reached yet.
In fact, while it would be tempting to consider all segmentation classes as equally important this could lead to penalizing excessively the score of a model that made some minor mistake on one of the classes that appear more sparsely in the dataset, which is not always the desired effect in a real-world application.
As expected the macro averaged scores tend to be much lower than their weighted average counterparts as typically the least represented classes are also the hardest ones to classify correctly.

As we can see the model that presents the overall best performance for both weighted and macro averages on the full dataset is DeepLabV3+\cite{deeplabv3plus}. The only exception is represented by the macro averaged recall, for which DeepLabV3\cite{deeplab} achieves the best results.

A possible reason behind this is that both systems introduce, as part of their architecture, a way of effectively leveraging the global context of the image instead of just relying on local information. In contrast, the FCN, which doesn't implement any kind of global information, shows a much lower performance on the task at hand.
While LRSAPP and PSPNet also rely on global information to perform the segmentation, they likely need a much larger amount of data to be properly trained and reach convergence.

In order to provide a more in-depth insight into the results obtained on the presented dataset we provide a visualization of the confusion matrices obtained by applying the best performing model, namely DeepLabV3+, on each of the manuscript classes that make up U-DIADS-Bib (Fig. \ref{fig:confmatrix}). We can notice how each of the manuscripts presents different criticalities, in terms of segmentation classes that are particularly difficult to identify correctly. In particular, DeepLabV3+ obtained particularly low accuracies for the Paratext and Title classes on the "Latin 14396" and "Syriaque 341" manuscripts respectively, while both classes represented a challenge when working with the "Latin 2" manuscript. The performance on the "Latin 16746" manuscript, on the other hand, appears to be consistent for all the segmentation classes, with only a small drop in accuracy for the Title.


\paragraph{Few-shot setting}
As previously stated, as part of the present work we wanted to emphasize the importance of being able to work with a small amount of data while retaining good performances for the task of document layout analysis. For this reason, in Table~\ref{tab:fewshotresult}, we also present the results obtained by the different segmentation approaches on U-DIADS-BibFS, the few-shot version of the proposed dataset. 
As previously described, only three images were adopted for the training process. Consistently with the full version of the dataset, the results are provided both for a single manuscript and for the full dataset, and for weighted and macro averages.

In this setting, the De Nardin et al.\cite{WACV}, which represents the current state-of-the-art for few-shot document layout segmentation, achieved the best results for all the metrics, both  when using the weighted and macro averaging strategies on the full dataset, while being outperformed by either DeepLabV3 or DeepLabV3+ on some of the metrics when looking at the individual manuscripts scores.

\begin{figure*}[tb]%
    \centering
    \begin{adjustbox}{minipage=\linewidth,scale=0.8}
    \subfloat[\centering Latin 2]{{\includegraphics[width=.45\linewidth]{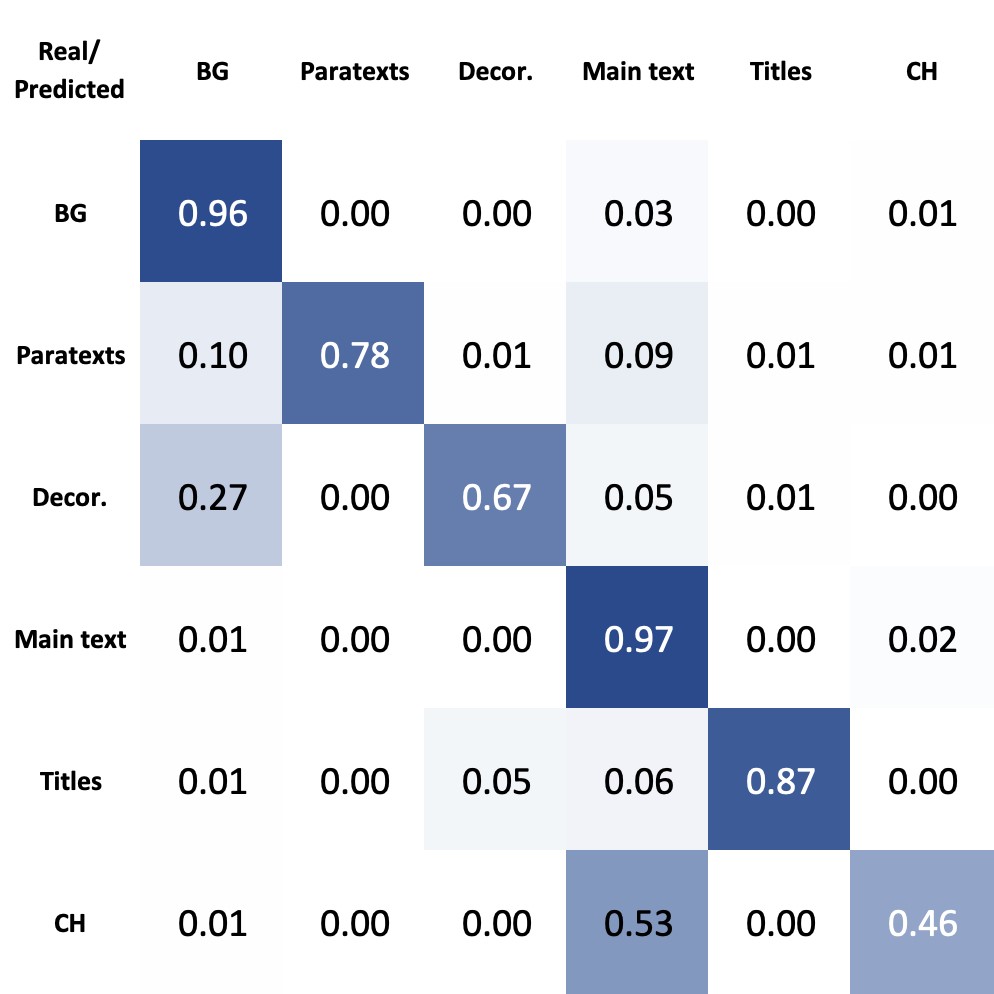} \label{fig:l2confFS} }}%
    \subfloat[\centering Latin 14396]{{\includegraphics[width=.45\linewidth]{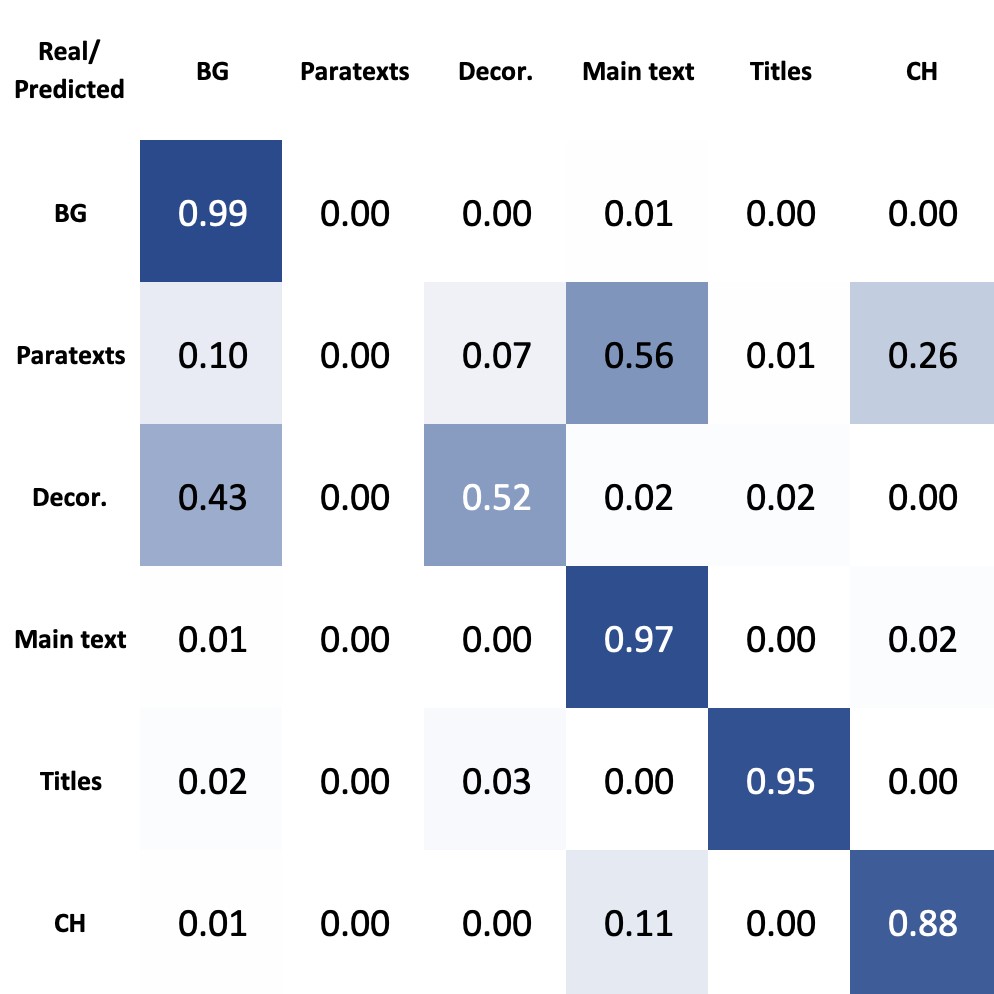} \label{fig:l14confFS}}}
    \\
    \subfloat[\centering Latin 16746]{{\includegraphics[width=.45\linewidth]{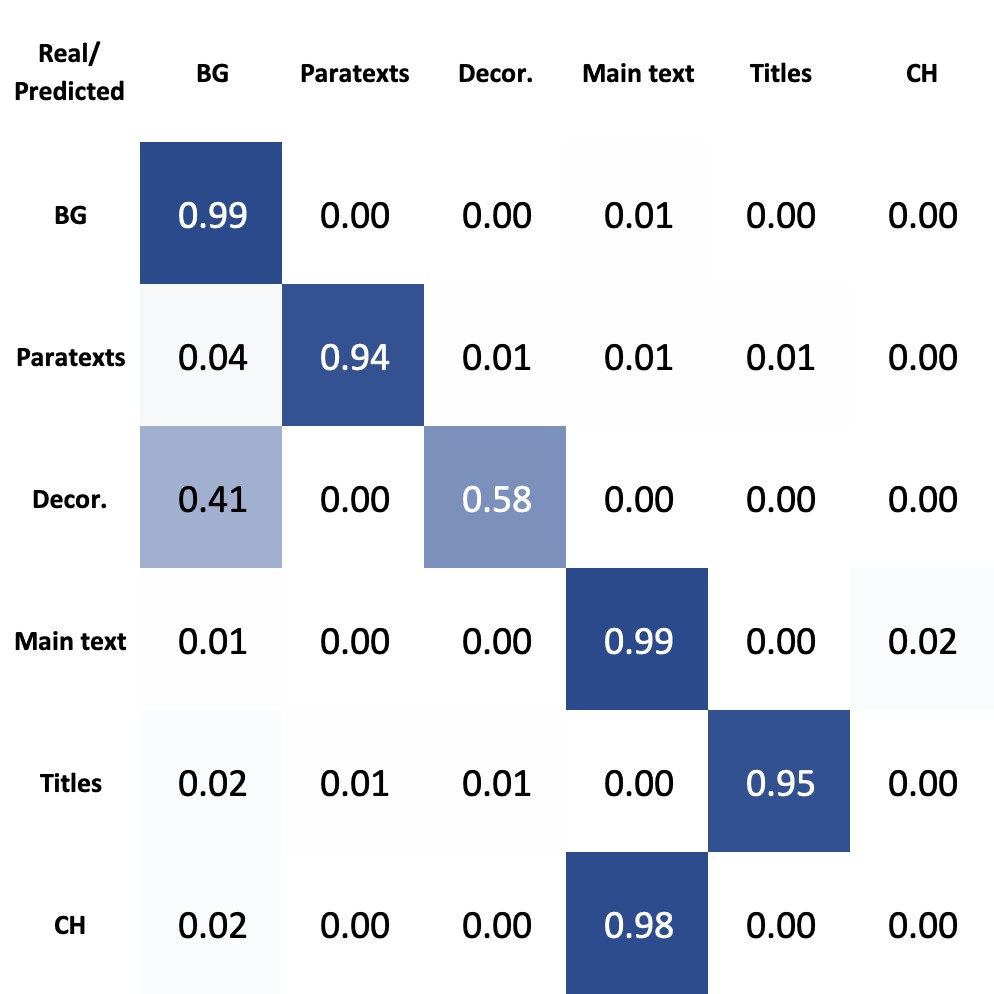} \label{fig:l16confFS} }}%
    \subfloat[\centering Syriaque 341]{{\includegraphics[width=.45\linewidth]{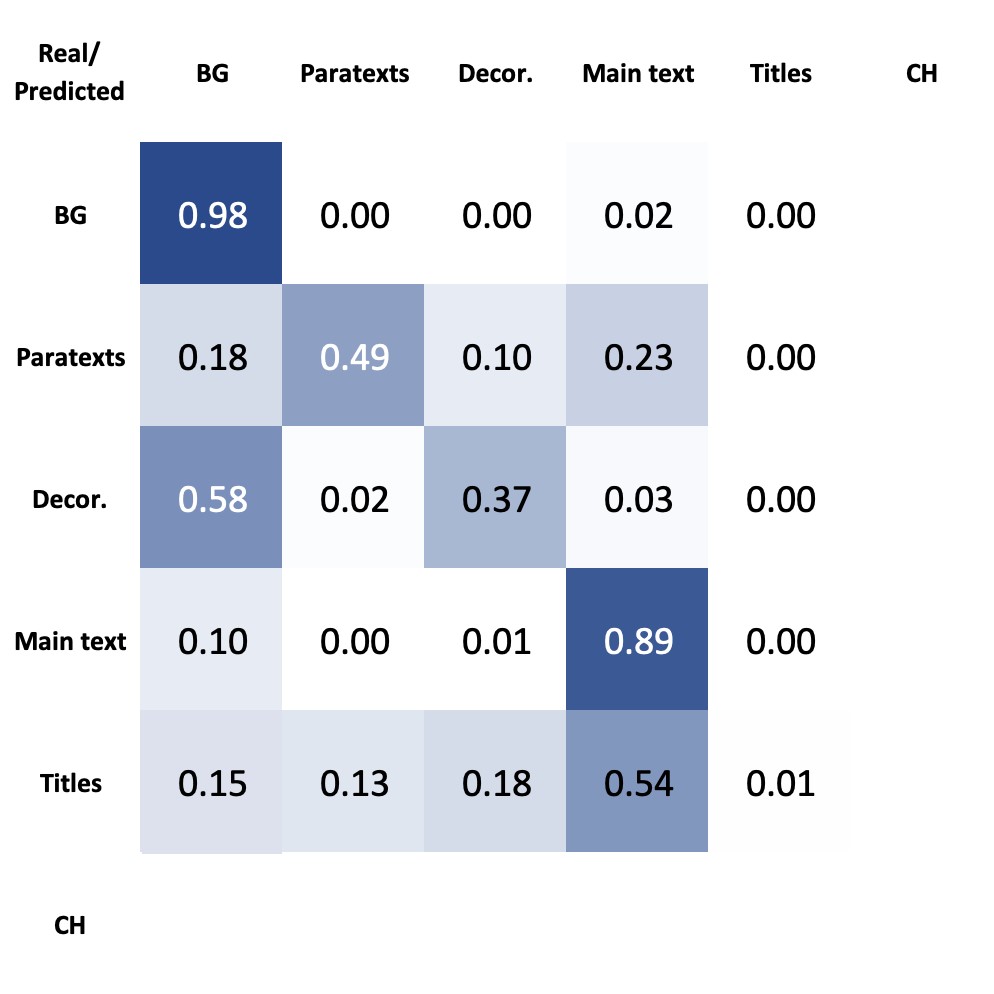} }}
    \caption{Confusion matrices obtained by applying  the model proposed by De Nardin et al.\cite{WACV} to the four ancient manuscripts of the U-DIADS-BibFS dataset.}
    \label{fig:confmatrixfs}%
    \end{adjustbox}
\end{figure*}

As we did for the full dataset, in Figure~\ref{fig:confmatrixfs} we provide the results obtained by the best-performing model on U-DIADS-BibFS. As expected, for most of the manuscripts we can observe a deterioration of the scores obtained for the different classes. In particular, we can notice substantial drop in accuracy for the Paratext and Decoration classes on the "Latin 14396" and "Syriaque 341" manuscripts. Similarly, the Decoration and CH classes of the "Latin 16746" manuscript appear to be much harder to identify in the few-shot learning setting compared to the full dataset one. Surprisingly, the performance on the "Latin 2" manuscript is overall better in this scenario, where we can notice a substantial improvement in accuracy for the Paratext, Main Text and CH classes respectively, counterbalanced only by a noticeable drop in accuracy for the Decoration class.

\section{Conclusion and Future work}\label{sec5}
This paper introduced U-DIADS-Bib, a novel dataset for layout analysis of handwritten documents,
which effectively addresses some of the major drawbacks of the other datasets currently available in the literature, mainly represented by lack of detail in the segmentation process or in a low quality of the provided segmentation ground truths. Both of these aspects are crucial in order to obtain reliable systems that can be employed in a real-world scenario.
As a further contribution, we have also alleviated the time and cost-consuming problem of manual annotation of GTs by expert humanists, by presenting a hybrid segmentation pipeline that allows experts to annotate large numbers of pages with the help of ML-based automated approaches.

From the presented benchmark emerged that, while the current models are able to effectively identify some of the layout classes for most of the manuscripts composing the dataset, other classes still pose a significant challenge, especially when working in the few-shot learning setting.
These criticalities are reflected by both the low accuracies for these classes in the confusion matrices of the different manuscripts and the low scores obtained by the selected segmentation approaches on the macro averaged metrics.
The proposed dataset is clearly characterized by a variety of challenging aspects, such as the close similarity between some of the layout classes and it highly imbalanced nature, which we believe have the potential to stimulate further improvements in this field of research especially when combined with the further challenge represented by low data availability of the U-DIADS-BibFS version.

As part of our future work we plan to expand the dataset by introducing other types of documents with different characteristics, in order to further expand its heterogeneity. Potentially interesting ideas would be to introduce modern manuscripts, for example in the form of notes or letter, as well as printed documents with manual annotations.
Also, a more refined classification of the paratexts by expanding the number of segmentation classes is certainly one of the future objectives of the project. We believe a hierarchical representation of the layout structure could represent an interesting step forward for future versions of the dataset.
Finally, we plan to provide the GTs also for the text-line segmentation and baseline detection tasks, in order to provide a more complete dataset, capable of addressing all the different tasks connected to document layout analysis.

On the application side, one the other hand, an interesting idea for future works could be represented by the introduction of modules focused on specific subsets of features (such as colour \cite{AMELIOcolor}) as a way to further improve the performance of the current segmentation architectures.
Furthermore, it could be worthwhile to explore the adoption of techniques such as knowledge distillation \cite{hinton2015distilling} or network pruning \cite{AMELIOcompression} to obtain networks characterized by a more compact structure.

\backmatter
\bmhead{Acknowledgments}
The authors would like to acknowledge the Bibliothèque nationale de France for providing access to the digital library Gallica.

\bmhead{Data Availability}
The datasets generated and analysed during the current study are available in the U-DIADS-Bib repository\footnotemark[7].
\footnotetext[7]{https://ai4ch.uniud.it/udiadsbib/}

\section*{Declarations}
\bmhead{Funding} Partial financial support was received from Piano Nazionale di Ripresa e Resilienza (PNRR) DD 3277 del 30 dicembre 2021 (PNRR Missione 4, Componente 2, Investimento 1.5) - Interconnected Nord-Est Innovation Ecosystem (iNEST).

\bmhead{Conflict of interest} The authors declare no conflict of interest.

\bibliographystyle{sn-basic}
\bibliography{sn-bibliography.bib}

\begin{thebibliography}{34}
\providecommand{\natexlab}[1]{#1}
\providecommand{\url}[1]{{#1}}
\providecommand{\urlprefix}{URL }
\providecommand{\doi}[1]{\url{https://doi.org/#1}}
\providecommand{\eprint}[2][]{\url{#2}}
 \bibcommenthead

\bibitem[{Adam et~al(2018)Adam, Baig, Al-Maadeed, Bouridane, and
  El-Menshawy}]{Adam2018}
Adam K, Baig A, Al-Maadeed S, et~al (2018) Kertas: dataset for automatic dating
  of ancient arabic manuscripts. International Journal on Document Analysis and
  Recognition (IJDAR) 21(4):283--290. \doi{10.1007/s10032-018-0312-3}

\bibitem[{Alaei et~al(2011)Alaei, Nagabhushan, and Pal}]{phtd}
Alaei A, Nagabhushan P, Pal U (2011) A new dataset of persian handwritten
  documents and its segmentation. In: 2011 7th Iranian Conference on Machine
  Vision and Image Processing, pp 1--5, \doi{10.1109/IranianMVIP.2011.6121553}

\bibitem[{Amelio et~al(2022)Amelio, Bonifazi, Corradini, {Di Saverio},
  Marchetti, Ursino, and Virgili}]{AMELIOcolor}
Amelio A, Bonifazi G, Corradini E, et~al (2022) Defining a deep neural network
  ensemble for identifying fabric colors. Applied Soft Computing 130:109,687.
  \doi{https://doi.org/10.1016/j.asoc.2022.109687}

\bibitem[{Amelio et~al(2023)Amelio, Bonifazi, Cauteruccio, Corradini,
  Marchetti, Ursino, and Virgili}]{AMELIOcompression}
Amelio A, Bonifazi G, Cauteruccio F, et~al (2023) Representation and
  compression of residual neural networks through a multilayer network based
  approach. Expert Systems with Applications 215:119,391.
  \doi{https://doi.org/10.1016/j.eswa.2022.119391}

\bibitem[{Boillet et~al(2019)Boillet, Bonhomme, Stutzmann, and
  Kermorvant}]{horae}
Boillet M, Bonhomme ML, Stutzmann D, et~al (2019) Horae: An annotated dataset
  of books of hours. In: Proceedings of the 5th International Workshop on
  Historical Document Imaging and Processing. Association for Computing
  Machinery, New York, NY, USA, HIP '19, p 7–12,
  \doi{10.1145/3352631.3352633}

\bibitem[{Bukhari et~al(2012)Bukhari, Breuel, Asi, and El-Sana}]{Bukhari}
Bukhari SS, Breuel TM, Asi A, et~al (2012) Layout analysis for arabic
  historical document images using machine learning. In: 2012 International
  Conference on Frontiers in Handwriting Recognition, pp 639--644,
  \doi{10.1109/ICFHR.2012.227}

\bibitem[{Chen et~al(2017)Chen, Papandreou, Schroff, and Adam}]{deeplab}
Chen L, Papandreou G, Schroff F, et~al (2017) Rethinking atrous convolution for
  semantic image segmentation. CoRR abs/1706.05587.
  {\href{https://arxiv.org/abs/1706.05587}{{https://arxiv.org/abs/1706.05587}}}

\bibitem[{Chen et~al(2018)Chen, Zhu, Papandreou, Schroff, and
  Adam}]{deeplabv3plus}
Chen LC, Zhu Y, Papandreou G, et~al (2018) Encoder-decoder with atrous
  separable convolution for semantic image segmentation. In: Ferrari V, Hebert
  M, Sminchisescu C, et~al (eds) Computer Vision -- ECCV 2018. Springer
  International Publishing, Cham, pp 833--851

\bibitem[{Cilia et~al(2021)Cilia, De~Stefano, Fontanella, Marthot-Santaniello,
  and Scotto~di Freca}]{Cilia}
Cilia ND, De~Stefano C, Fontanella F, et~al (2021) Papyrow: A dataset of row
  images from ancient greek papyri for writers identification. In: Del~Bimbo A,
  Cucchiara R, Sclaroff S, et~al (eds) Pattern Recognition. ICPR International
  Workshops and Challenges. Springer International Publishing, Cham, pp
  223--234

\bibitem[{Clausner et~al(2018)Clausner, Antonacopoulos, Mcgregor, and
  Wilson-Nunn}]{RASM2018}
Clausner C, Antonacopoulos A, Mcgregor N, et~al (2018) Icfhr 2018 competition
  on recognition of historical arabic scientific manuscripts – rasm2018. In:
  2018 16th International Conference on Frontiers in Handwriting Recognition
  (ICFHR), pp 471--476, \doi{10.1109/ICFHR-2018.2018.00088}

\bibitem[{De~Nardin et~al(2023{\natexlab{a}})De~Nardin, Zottin, Paier, Foresti,
  Colombi, and Piciarelli}]{WACV}
De~Nardin A, Zottin S, Paier M, et~al (2023{\natexlab{a}}) Efficient few-shot
  learning for pixel-precise handwritten document layout analysis. In: 2023
  IEEE/CVF Winter Conference on Applications of Computer Vision (WACV), pp
  3669--3677, \doi{10.1109/WACV56688.2023.00367}

\bibitem[{De~Nardin et~al(2023{\natexlab{b}})De~Nardin, Zottin, Piciarelli,
  Colombi, and Foresti}]{de2023few}
De~Nardin A, Zottin S, Piciarelli C, et~al (2023{\natexlab{b}}) Few-shot
  pixel-precise document layout segmentation via dynamic instance generation
  and local thresholding. International Journal of Neural Systems
  33(10):2350,052. \doi{10.1142/S0129065723500521}

\bibitem[{Dolfing et~al(2020)Dolfing, Bellegarda, Chorowski, Marxer, and
  Laurent}]{Dolfing}
Dolfing HJ, Bellegarda J, Chorowski J, et~al (2020) The “scribblelens”
  dutch historical handwriting corpus. In: 2020 17th International Conference
  on Frontiers in Handwriting Recognition (ICFHR), pp 67--72,
  \doi{10.1109/ICFHR2020.2020.00023}

\bibitem[{Fiel et~al(2017)Fiel, Kleber, Diem, Christlein, Louloudis, Nikos, and
  Gatos}]{Fiel}
Fiel S, Kleber F, Diem M, et~al (2017) Icdar2017 competition on historical
  document writer identification (historical-wi). In: 2017 14th IAPR
  International Conference on Document Analysis and Recognition (ICDAR), pp
  1377--1382, \doi{10.1109/ICDAR.2017.225}

\bibitem[{Fischer et~al(2010)Fischer, Inderm\"{u}hle, Bunke, Viehhauser, and
  Stolz}]{Fischer1}
Fischer A, Inderm\"{u}hle E, Bunke H, et~al (2010) Ground truth creation for
  handwriting recognition in historical documents. In: Proceedings of the 9th
  IAPR International Workshop on Document Analysis Systems. Association for
  Computing Machinery, New York, NY, USA, DAS '10, p 3–10,
  \doi{10.1145/1815330.1815331}

\bibitem[{Fischer et~al(2011)Fischer, Frinken, Forn\'{e}s, and Bunke}]{Fischer}
Fischer A, Frinken V, Forn\'{e}s A, et~al (2011) Transcription alignment of
  latin manuscripts using hidden markov models. In: Proceedings of the 2011
  Workshop on Historical Document Imaging and Processing. Association for
  Computing Machinery, New York, NY, USA, HIP '11, p 29–36,
  \doi{10.1145/2037342.2037348}

\bibitem[{Gatos et~al(2015)Gatos, Stamatopoulos, Louloudis, Sfikas, Retsinas,
  Papavassiliou, Sunistira, and Katsouros}]{Gatos}
Gatos B, Stamatopoulos N, Louloudis G, et~al (2015) Grpoly-db: An old greek
  polytonic document image database. In: 2015 13th International Conference on
  Document Analysis and Recognition (ICDAR), pp 646--650,
  \doi{10.1109/ICDAR.2015.7333841}

\bibitem[{Grüning et~al(2018)Grüning, Labahn, Diem, Kleber, and
  Fiel}]{Gruning}
Grüning T, Labahn R, Diem M, et~al (2018) Read-bad: A new dataset and
  evaluation scheme for baseline detection in archival documents. In: 2018 13th
  IAPR International Workshop on Document Analysis Systems (DAS), pp 351--356,
  \doi{10.1109/DAS.2018.38}

\bibitem[{Hinton et~al(2015)Hinton, Vinyals, and Dean}]{hinton2015distilling}
Hinton G, Vinyals O, Dean J (2015) Distilling the knowledge in a neural
  network. arXiv preprint arXiv:150302531

\bibitem[{Howard et~al(2019)Howard, Sandler, Chen, Wang, Chen, Tan, Chu,
  Vasudevan, Zhu, Pang, Adam, and Le}]{lraspp}
Howard A, Sandler M, Chen B, et~al (2019) Searching for mobilenetv3. In: 2019
  IEEE/CVF International Conference on Computer Vision (ICCV), pp 1314--1324,
  \doi{10.1109/ICCV.2019.00140}

\bibitem[{Kassis et~al(2017)Kassis, Abdalhaleem, Droby, Alaasam, and
  El-Sana}]{Kassis}
Kassis M, Abdalhaleem A, Droby A, et~al (2017) Vml-hd: The historical arabic
  documents dataset for recognition systems. In: 2017 1st International
  Workshop on Arabic Script Analysis and Recognition (ASAR), pp 11--14,
  \doi{10.1109/ASAR.2017.8067751}

\bibitem[{Kiessling et~al(2019)Kiessling, Ezra, and Miller}]{Kiessling}
Kiessling B, Ezra DSB, Miller MT (2019) Badam: A public dataset for baseline
  detection in arabic-script manuscripts. In: Proceedings of the 5th
  International Workshop on Historical Document Imaging and Processing.
  Association for Computing Machinery, New York, NY, USA, HIP '19, p 13–18,
  \doi{10.1145/3352631.3352648}

\bibitem[{Kurar~Barakat et~al(2019)Kurar~Barakat, El-Sana, and Rabaev}]{Pinkas}
Kurar~Barakat B, El-Sana J, Rabaev I (2019) The pinkas dataset. In: 2019
  International Conference on Document Analysis and Recognition (ICDAR), pp
  732--737, \doi{10.1109/ICDAR.2019.00122}

\bibitem[{Long et~al(2015)Long, Shelhamer, and Darrell}]{fcn}
Long J, Shelhamer E, Darrell T (2015) Fully convolutional networks for semantic
  segmentation. In: 2015 IEEE Conference on Computer Vision and Pattern
  Recognition (CVPR), pp 3431--3440, \doi{10.1109/CVPR.2015.7298965}

\bibitem[{Mehri et~al(2017)Mehri, H\'{e}roux, Mullot, Moreux, Co\"{u}asnon, and
  Barrett}]{HBA}
Mehri M, H\'{e}roux P, Mullot R, et~al (2017) Hba 1.0: A pixel-based annotated
  dataset for historical book analysis. In: Proceedings of the 4th
  International Workshop on Historical Document Imaging and Processing.
  Association for Computing Machinery, New York, NY, USA, HIP2017, p 107–112,
  \doi{10.1145/3151509.3151528}

\bibitem[{Nikolaidou et~al(2022)Nikolaidou, Seuret, Mokayed, and
  Liwicki}]{Nikolaidou2022}
Nikolaidou K, Seuret M, Mokayed H, et~al (2022) A survey of historical document
  image datasets. International Journal on Document Analysis and Recognition
  (IJDAR) 25(4):305--338. \doi{10.1007/s10032-022-00405-8}

\bibitem[{Potanin et~al(2021)Potanin, Dimitrov, Shonenkov, Bataev, Karachev,
  Novopoltsev, and Chertok}]{Potanin}
Potanin M, Dimitrov D, Shonenkov A, et~al (2021) Digital peter: New dataset,
  competition and handwriting recognition methods. In: The 6th International
  Workshop on Historical Document Imaging and Processing. Association for
  Computing Machinery, New York, NY, USA, HIP '21, p 43–48,
  \doi{10.1145/3476887.3476892}

\bibitem[{Quirós et~al(2020)Quirós, Kallio, and Vidal}]{quiros}
Quirós L, Kallio M, Vidal E (2020) {Finnish Court Records-sub500. A dataset of
  Finnish notarial records (19th Century)}. \doi{10.5281/zenodo.3945088}

\bibitem[{Romero and Sánchez(2021)}]{hisclima}
Romero V, Sánchez JA (2021) The hisclima database: historical weather logs for
  automatic transcription and information extraction. In: 2020 25th
  International Conference on Pattern Recognition (ICPR), pp 10,141--10,148,
  \doi{10.1109/ICPR48806.2021.9412210}

\bibitem[{Saini et~al(2019)Saini, Dobson, Morrey, Liwicki, and
  Simistira~Liwicki}]{Chinese}
Saini R, Dobson D, Morrey J, et~al (2019) Icdar 2019 historical document
  reading challenge on large structured chinese family records. In: 2019
  International Conference on Document Analysis and Recognition (ICDAR), pp
  1499--1504, \doi{10.1109/ICDAR.2019.00241}

\bibitem[{Sauvola and Pietikäinen(2000)}]{SAUVOLA}
Sauvola J, Pietikäinen M (2000) Adaptive document image binarization. Pattern
  Recognition 33(2):225--236. \doi{10.1016/S0031-3203(99)00055-2}

\bibitem[{Simistira et~al(2016)Simistira, Seuret, Eichenberger, Garz, Liwicki,
  and Ingold}]{diva}
Simistira F, Seuret M, Eichenberger N, et~al (2016) Diva-hisdb: A precisely
  annotated large dataset of challenging medieval manuscripts. In: 2016 15th
  International Conference on Frontiers in Handwriting Recognition (ICFHR), pp
  471--476, \doi{10.1109/ICFHR.2016.0093}

\bibitem[{Wüthrich et~al(2009)Wüthrich, Liwicki, Fischer, Indermühle, Bunke,
  Viehhauser, and Stolz}]{Wüthrich}
Wüthrich M, Liwicki M, Fischer A, et~al (2009) Language model integration for
  the recognition of handwritten medieval documents. In: 2009 10th
  International Conference on Document Analysis and Recognition, pp 211--215,
  \doi{10.1109/ICDAR.2009.17}

\bibitem[{Zhao et~al(2017)Zhao, Shi, Qi, Wang, and Jia}]{pspnet}
Zhao H, Shi J, Qi X, et~al (2017) Pyramid scene parsing network. In: 2017 IEEE
  Conference on Computer Vision and Pattern Recognition (CVPR), pp 6230--6239,
  \doi{10.1109/CVPR.2017.660}

\end{thebibliography}


\end{document}